\def\something{} 
\ifdefined\something
\documentclass[1p,a4,10pt]{elsarticle}
\let\today\relax
\makeatletter
\def\ps@pprintTitle{%
    \let\@oddhead\@empty
    \let\@evenhead\@empty
    \def\@oddfoot{\footnotesize\itshape
         {} \hfill\today}
    \let\@evenfoot\@oddfoot
    }
\makeatother
\else
\documentclass[10pt,journal,compsoc]{IEEEtran}
\fi
\usepackage[cmex10]{amsmath}
\usepackage[tight,footnotesize]{subfigure}

\usepackage[utf8]{inputenc} 
\usepackage[T1]{fontenc}    
\usepackage{hyperref}       
\usepackage{url}            
\usepackage{xurl}
\usepackage{booktabs}       
\usepackage{amsfonts}       
\usepackage{nicefrac}       
\usepackage{microtype}      
\usepackage{url}

\usepackage{graphics}
\usepackage{tabularx}
\usepackage{multirow}
\usepackage{booktabs}
\usepackage{graphicx}
\usepackage{epsfig}
\usepackage{mathrsfs}
\usepackage{color}
\usepackage{amssymb}
\usepackage{amsmath}
\usepackage{amsthm}
\usepackage{algorithm}
\usepackage{algpseudocode}
\usepackage{subfigure}
\usepackage{euscript}
\usepackage{eufrak} 
\usepackage{bm}
\usepackage{verbatim}
\usepackage[normalem]{ulem}
\usepackage{xcolor}
\usepackage{comment}
\usepackage{filecontents}
\usepackage{silence}
\newcommand*\rot{\rotatebox{90}}
\def\mkcolorsup#1#2{\mbox{\small{\color{#1}$^\text{\rm #2}$}}}

\newcommand{\ie}{\textit{i.e.}}
\newcommand{\eg}{\textit{e.g.}}

\usepackage{color}
\usepackage{epsfig} 
\graphicspath{{.}{./figs/}}
\DeclareGraphicsExtensions{.eps}

\DeclareMathOperator*{\argmax}{arg\,max}
\DeclareMathOperator*{\argmin}{arg\,min}
\def\req#1{(\ref{#1})}

\def\mathcal#1{\mbox{$\cal{#1}$}}

\def\req#1{(\ref{#1})}

\newtheorem{mystat}{Statement}

\newcommand\blfootnote[1]{%
  \begingroup
  \renewcommand\thefootnote{}\footnote{#1}%
  \addtocounter{footnote}{-1}%
  \endgroup
}

\ifdefined\something
\journal{Pattern Recognition}
\fi
 
\begin{document}
\def\abstractt{
This paper presents a novel feature selection method based on the conditional mutual information (CMI). The proposed High Order Conditional Mutual Information Maximization (HOCMIM) incorporates high order dependencies into the feature selection procedure and has a straightforward interpretation due to its bottom-up derivation. The HOCMIM is derived from the CMI's chain expansion and expressed as a maximization optimization problem. The maximization problem is solved using a greedy search procedure, which speeds up the entire feature selection process. The experiments are run on a set of benchmark datasets (20 in total). The HOCMIM is compared with eighteen state-of-the-art feature selection algorithms, from the results of two supervised learning classifiers (Support Vector Machine and K-Nearest Neighbor). The HOCMIM achieves the best results in terms of accuracy and shows to be faster than high order feature selection counterparts.
}

\def\titulo{High-Order Conditional Mutual Information Maximization for dealing with High-Order Dependencies in Feature Selection\blfootnote{{\em Paper accepted at Pattern Recognition Journal}: \url{https://doi.org/10.1016/j.patcog.2022.108895}}}

\ifdefined\something
\begin{frontmatter}
\title{\titulo}
\author[label1]{Francisco Souza}
\ead{fasouza@isr.uc.pt}
\author[label1]{Cristiano Premebida}
\ead{cpremebida@isr.uc.pt}
\author[label1]{Rui Araújo}
\ead{rui@isr.uc.pt}

\address[label1]{University of Coimbra, Institute of Systems and Robotics, Portugal.
}

\begin{abstract}
\abstractt
\end{abstract}

\end{frontmatter}

\else
\title{\titulo}
\author{Francisco~Souza, Cristiano~Premebida, Rui~Araújo
\thanks{F.Souza is with Radboud University, the Netherlands.
C.Premebida and R.Ara\'ujo are with the University of Coimbra, Institute of Systems and Robotics, Department of Electrical and Computer Engineering, Portugal. Emails: \footnotesize\{fasouza,~cpremebida,~rui\}@isr.uc.pt.}
}

\maketitle
\begin{abstract}
\abstractt
\end{abstract}

\IEEEpeerreviewmaketitle

\fi
\section{Introduction}
\label{sec:intro}
Feature selection (FS) plays a vital role in machine learning (ML), with the objective being to select a relevant (and small) subset from the complete (and large) set of features. The FS can benefit the ML models by allowing the learning of less complex models, reducing the risk of overfitting, allowing the use of less memory for data storage and, helping with the model interpretability \cite{Guyon2003}. FS techniques aim to remove redundant and irrelevant features, and they are classified on three distinct methods \cite{Guyon2003}: wrapper, embedded, {or} filter {methods}. In wrapper methods, the subset selection is driven by the model performance \ie, the objective is to select a subset that leads to the highest model accuracy. The wrapper requires training the model every time a subset is evaluated, being unsuitable for high dimensional data due to the computational requirements. Embedded methods use intrinsic characteristics of the model as a basis for computing the quality of a subset (\eg, sensitivity analysis) or by selecting a subset along with the model learning (\eg, regularization), while being computationally faster than wrapper methods. On the other hand, filter methods rely on statistical measures to compute the quality of a subset. Filter methods are also known as model-free techniques (\ie, model agnostic) and {are faster than} wrapper and embedded {methods}. 

{\color{black}
There are a variety of methods for FS in ML, most of which are based on embedded methods and filter methods. The LASSO $\ell_1$ regularization penalty \cite{Tibshirani96}, and the recursive feature elimination (RFE) \cite{Guyon2003}, are both well-known embedded FS approaches. More recently, the authors in  \cite{Peng2021} proposed the discriminative ridge machine (DRM) that allows to deal with classification for high dimensional and imbalanced data. The authors of \cite{Jinglin2020} expanded the SVM for multi-view learning applications, and they incorporated a FS stage to be to select important and discriminative features. By using regularization, the authors in \cite{Zhu2015} and \cite{Tianji2019} proposed to employ $\ell_{p,(0 < p \leq 1)}$ norm and $\ell_{2,0}$ norm for feature selection. In \cite{Zhu2015}, the authors proposed an accelerated robust subset selection method based on the $\ell_{p,(0 < p \leq 1)}\text{-norm}$. In \cite{Tianji2019} the authors propose a novel framework to solve the $\ell_{2,0}$ constrained multi-class FS problem, by representing the problem into the discriminant analysis (LDA) and solving the constrain under this prism. 

Embedded methods are still model dependent, on the other hand, filter methods are model agnostic, and are applicable when the type of classifiers has not been decided yet. In \cite{Chao2017}, the authors proposed a unsupervised feature selection method based on the Locally Linear Embedding (LLE) score as a statistical measure of feature relevance. The infinite latent feature selection (ILFS) is proposed in \cite{Roffo2017}. The features are ranked based on their similarity to their neighbors using causal graphs. The Neighborhood component analysis (NCA) \cite{NCA} is a non-parametric FS algorithm based on the $k$-nearest neighbors ($k$NN) that employs a gradient ascent technique to maximize the expected leave-one-out classification accuracy with a regularization term. The Relieff is a filter approach that ranks the features based on a score of sample proximity \cite{liu2008}. The authors of \cite{Zhang2022} propose a novel feature ranking criterion based on the squared orthogonal correlation coefficient, known as orthogonal least squares FS (OLS-FS). The canonical correlation coefficient and linear discriminant analysis are linked to the OLS-FS ranking criteria. 
%
Distance scores, such as the Fischer score, Gini index, or Kruskal-Wallis test \cite{Kruska}, are also frequently used as FS filters. The preceding methods are parametric, assuming that the data follows a parametric distribution. Non parametric distance scores such as belief entropy \cite{Xiao2020}, divergence entropy \cite{Xiao2020b}, and mutual information \cite{Battiti94} can also be used for FS.
}
This paper focuses on filter methods relying on information theory, specifically on mutual information (MI). There are several advantages of using MI for feature selection, namely: linear and non-linear {dependencies} can be captured, invariance under feature transformation, {suitability both to} classification and regression, {\color{black}and it can handle imbalanced data}.

The MI-based feature selection is widely applied in many domains, such {as genomics}, chemometrics, industry, robotics, health, {and} remote sensing.
The applicability of MI as a criterion for feature selection has started on the seminal work of Batitti et al.\ \cite{Battiti94}, where the method called Mutual Information-Based Feature Selection (MIFS) was proposed. {In MIFS}, one feature is added sequentially to the set of selected features, and, the selected feature is the one that has {the} most relevance with respect to the target and less redundancy w.r.t.\ the set of selected features. The MI between the candidate feature and the target is used to determine relevance, while the average MI between the candidate feature and the set of selected features is used to determine redundancy. Since then, several other approaches based on the concepts of ``relevance'' and ``redundancy'' have been proposed, such as the well known minimum Redundancy Maximum Relevance (mRMR) algorithm \cite{Hanchuan2005} and other variants \cite{Kwak2002}, \cite{Pablo2009}. {\color{black} In \cite{Kwak2002} the authors approximate the MI by considering feature interaction. In \cite{Pablo2009} the authors employed the normalization of the redundancy term.} {Essentially, these} methods make an approximation of high dimensional MI from second-order estimators\cite{Balagani2010}. 

In Brown et al.\ \cite{Brown2012} the authors demonstrated the link between these MI-FS and the maximum likelihood, stating that maximization of CMI in an FS perspective is equivalent to maximization of likelihood. The authors then explored the relationship between second-order estimators and the CMI. In most cases, such methods rely on CMI approximations that are second-order order and non-linear. As addition to the previous methods, there are the Conditional Mutual Information Maximization (CMIM) criterion \cite{Fleuret2004}, the Joint Mutual Information (JMI) \cite{Yang1999}, the Double Input Symmetrical relevance (DISR) \cite{Meyer2008}, and variants of previous methods, such as the {\color{black} lower bound of CMI (OLB-CM) \cite{Peng2017}, and the joint mutual information maximisation (JMIM) \cite{Bennasar2015}}.
Other relevant works solve the MI-FS as an optimization problem \cite{Nguyen2014}, {focus} on improving the MI estimators \cite{Sharmin2019,Sechidis2019}, and work by selecting the stopping rules for FS \cite{Mielniczuk2019}.



{
High-order MI FS are more recent in the literature. The Relax-mRMR \cite{Xuan2016}, inspired by the mRMR algorithm, was designed to handle $3^{rd}$ order dependencies, but at a high computational cost. Another approach is the {Conditional Mutual Information with Complementary and Opposing Teams (CMICOT)} algorithm proposed in \cite{Shishkin2016} that is generic and can explores any order dependencies; however, despite the clear motivation from the learning of decision trees ensembles, the connection with the CMI is not straightforward. In \cite{Sechidis2019}, high order expansions for the CMIM and JMI were derived by considering $3^{rd}$ and $4^{th}$ order interactions. These estimators were based on a series of ad-hoc assumptions and with no direct link to the CMI. Also, the estimators in \cite{Sechidis2019} run at a high computational cost, which grows exponentially with the order dependency. {For} a complete review on MI-FS algorithms {the} paper \cite{Li2017} {is suggested}. 

{\color{black}One of the major challenges in the literature is that, despite the significant number of FS proposals that estimate the CMI by second-order dependencies estimators, there is a lack of methodologies exploring high-order dependencies\footnote{here, high order dependencies are defined as those with orders higher or equal than three.}}
{\color{black}
{This paper proposes} a novel FS algorithm aiming to estimate the CMI {both taking into consideration} higher-order dependencies {and} being faster than the other equivalent higher-order MI 
methods. The proposed FS method, called high order CMI maximization (HOCMIM), is derived in a bottom-up approach and has interesting properties: (i) it uses the expansion property of the CMI, which has a clear and straightforward interpretation, giving a new perspective on the high order FS problem; (ii) it converges to the true value of CMI as $n\rightarrow{D}$, where $n$ is the order of the dependencies and $D$ is the total number of features; (iii) the HOCMIM scales linearly with the order dependency, being faster than the state-of-the-art algorithms; (iv) present an algorithm solution for determining the order $n$ adaptively. The HOCMIM is evaluated over $20$ benchmark datasets, and compared with over $18$ state-of-art FS methods, namely the MI based, such as the CMIM \cite{Fleuret2004}, JMI \cite{Yang1999}, DISR \cite{Meyer2008}, mRMR \cite{Hanchuan2005}, Relax-mRMR \cite{Xuan2016}, CMICOT \cite{Shishkin2016}, {CMIM-3, CMIM-4, JMI-3 and JMI-4} \cite{Sechidis2019}, and the following filter methods: Gini index, Fisher score, Kruskal-Wallis test \cite{Kruska}, ILFS \cite{Roffo2017}, NCA \cite{NCA}, Relieff \cite{liu2008}, and the FS-OLS \cite{Zhang2022}.}


The paper is organized as follows. Section \ref{s.pre} starts with the notations and a brief review of MI and CMI. Setion \ref{sec.background} discusses the background and related work.  Section \ref{sec.method} presents the proposed HOCMIM method. Section \ref{sec.experiments} discusses the results on numerous benchmark datasets using k-NN and SVM supervised classifiers. Finally, section \ref{sec.conclusion} gives the concluding remarks.

\section{Review of Mutual Information}
\label{s.pre}

In this section, the preliminaries to support the remaining of the manuscript are given, including the notation, background and related work.


\subsection{Notation}
Discrete finite random variables are represented by capital letters and their values by the corresponding lowercase letters \eg, a given random variable $A$ has its corresponding values denoted by $a$. 
The discrete input and output (target) features are defined as $X=\{X_1, \ldots, X_d\} \subset {\mathbb{R}^D}$ and $Y\in \{1,\ldots,C\}$, respectively, where {$\mathbb{R}$ is the set of real numbers}, and $C$ {is} the total number of classes. 
The probability that a random variable $A$ takes a value $a$ \ie, $P(A=a)$, is denoted by its probability mass function (pmf)
as $p_A(a)$ or, simplifying the notation, by $p(a)$.

\subsection{Mutual Information}
\label{ss.mi}

To define MI, the concept of entropy will be introduced first. 
The entropy of a set of discrete random variables $X=\{X_1, \ldots, X_d\}$, also referred as Shannon's entropy (\cite{Ash1990}, Chapter 1), is defined by
\begin{align}
 H(X) &{= -\sum_{x \in X}^{} p(x) \log p(x),} \label{equ.ent_mul} \\
   &= -\sum_{x_1 \in X_j}^{} \ldots \sum_{x_d \in X_d}^{} p(x_{1},\ldots,x_{D}) \log p(x_{1},\ldots,x_{D}),
   \nonumber
\end{align}
where $x=[x_1,\ldots,x_d]^T$. In information theory, the entropy {$H(X)$} measures the uncertainty associated {with feature $X$}, larger is the entropy, more uncertain $X$ is. Another useful quantity, is the conditional uncertainty of {$Y$} given that {$X=x$}, which is defined as
\begin{align}
 {H(Y|X=x) = - \sum_{y \in Y}^{} p(y|X=x) \log p(y|X=x).}
\label{equ.cond_uni}
\end{align}
By conditioning on all values of $X$, the conditional entropy $H(Y|X)$ is then defined as:
{
\begin{align}
 H(Y|X) & = H(Y,X) - H(X), \nonumber\\
 &  = \sum_{x \in X}^{} p(x) H(Y|X=x) =-\sum_{x \in X}^{} \sum_{y \in Y}^{} p(x,y) \log p(y|x).
\label{equ.cond_full}
\end{align}
}\relax
It measures the uncertainty (modeled by entropy) of $Y$ given the information on $X$. The larger is the information that {$X$} has on $Y$, the smaller is the conditional entropy {$H(Y|X)$}, in the case $H(Y|X)=0$, $X$ contains all information on $Y$. In the case $Y$ and $X$ are independent, then ${H(Y|X)}=H(Y)$. The conditional entropy has bounds $0\leq H(Y|X)\leq H(Y)$.

The MI is a general measure of dependency between random variables. The MI between a variable {$X$} and $Y$ is defined as:
{
\begin{align}
I(X;Y) &=  -\sum_{x \in X}^{} \sum_{y \in Y}^{} p(x,y) \log \frac{p(x,y)}{p(x)p(y)},\nonumber \\
& =H(Y) - H(Y|X) = H(X) - H(X|Y),\nonumber\\
& =H(Y) + H(X)  - H(Y,X).
\label{equ.mi_cond}
\end{align}
}\relax
{When $X$} contains all the information on $Y$ \ie, {$H(Y|X)=0$}, {then $I(X;Y)=H(Y)$}. Also, if {$X$} and $Y$ are independent \ie, {$H(Y|X)=H(Y)$} then {$I(X;Y)=0$}. The MI is always non-negative, i.e. its lower bounded by zero $I(X;Y)\geqslant 0$. 

Another useful quantity is the conditional MI (CMI) $I(X;Y|Z)$, which measures the gain of information of $X$ over $Y$ when $Z$ is known, and it is defined as
\begin{align}
 I(X;Y|Z) &= I(X,Z;Y)-I(Z;Y)\nonumber \\
 &= I(Y,Z;X)-I(Z;X)
\label{equ:gain_mi}
\end{align}
If $X$ does not increase the information on $Y$, then $I(X;Y|Z)=0$, while the larger $I(X;Y|Z)$ is, the larger will be the contribution of $X$ to reduce the uncertainty on $Y$. The CMI can also be expressed in the entropy form
\begin{align}
 I(X;Y|Z) &= H(Y;Z)-H(Y;X,Z)\nonumber\\
 &=H(X;Z)-H(X;Y,Z)
 \label{equ:entropy_cmi}
\end{align}


\section{Background and Related Work}
\label{sec.background}
Most MI feature selection methods use the sequential forward search (SFS) as the search subset procedure, as SFS produces acceptable results and is computationally efficient, with a good balance between accuracy and speed. The method proposed herein and related work will focus on techniques that work with the SFS. Also, SFS is considered sufficient for the goals of this work.

In the SFS, one feature $X_k$ is added sequentially to the pool of selected features $S$. In the MI view, the conditional mutual information (CMI) is an appropriate score function in the SFS to quantify the quality of feature $X_k$ \cite{Brown2012}. The CMI quantifies the increase of information about $Y$ when adding a feature $X_k$ to an existing set $S$. It is defined as 
\begin{align}
 I(X_k;Y|S)&= I(X_k;Y) - I(X_k;S) + I(X_k;S|Y).
\label{equ.J_CMI}
\end{align}
The score function of the SFS when using the CMI, is then defined as $J^{\text{CMI}}(X_k) = I(X_k;Y|S)$. The relevance term $I(X_k;Y)$ measures the amount of information that variable $X_k$ shares to the target $Y$. The higher the relevance, the higher is the correlation between $X_k$ and $Y$. One might think that selecting only trough the relevance criteria (aka. maximum information maximization (MIM)) would lead to a relevant set of features to predict $Y$. However, this is not true, as this criteria would select also redundant features. In the CMI, the redundancy term $I(X_k;S)$ balances the relevance by measuring the amount of information shared between $X_k$ and the subset of selected features $S$; the redundancy has negative sign, and decreases the CMI as high it is. The last term, the conditional redundancy $I(X_k;S|Y)$ quantifies the redundancy between $X_k$ and $S$ when the information on $Y$ is known. The conditional redundancy has positive sign and increases the CMI as higher is the shared information between $X_k$, and $S$, when $Y$ is known.

The CMI is the quantity of interest to estimate in the SFS-FS methods. Different methods have been proposed in the literature to calculate the CMI based on low and high-order dependencies estimators. These methods are reviewed in the next section.

\subsection{Low order estimators}
The estimation of CMI Eq. \req{equ.J_CMI} can follow by considering only second order dependencies between the features; this is stated here as low order estimators. 
The low-order estimators are based on the following approximation of CMI:
\begin{align}
 &J^{\beta,\gamma}(X_k) =I(X_k;Y) - \beta\sum_{X_j \in S_{}} I(X_j;X_k) + \gamma \sum_{X_j \in S_{}}I(X_j;X_k|Y),
 \label{e.aprox.general}
\end{align}
\begin{sloppypar}
where the different methods differ in the values of $\gamma$ and $\beta$. For example, in mRMR \cite{Hanchuan2005}, $J^{\text{mRMR}}(X_k)=J^{(\beta=|S|^{-1},\gamma=0)}(X_k)$, in MIFS \cite{Battiti94}, $J^{\text{MIFS}}(X_k)=J^{(\beta\in[0,1],\gamma=0)}(X_k)$, and JMI \cite{Yang1999}, $J^{\text{JMI}}(X_k)=J^{(\beta=|S|^{-1},\gamma=|S|^{-1})}(X_k)$. The JMI is the only which takes the conditional redundancy into account.
\end{sloppypar}
The double input symmetrical relevance {method DISR \cite{Meyer2008}} is based on a modification of JMI, by applying the following normalization factor
\begin{align}
J^{\text{DISR}}(X_k) &= \sum_{X_j \in S} \frac{I(X_k,X_j;Y)}{H(X_k,X_j,Y)}.
\end{align}
However, there is no theoretical justification for this normalization factor, as pointed out in \cite{Brown2012}. Another low order approach is the (conditional mutual information maximization) CMIM criteria, proposed in \cite{Fleuret2004}
\begin{align}
J^{\text{CMIM}}(X_k) &= \min_{X_j \in S} I(X_k;Y|X_j),
\end{align}
This is a nonlinear approximation to CMI, with no mathematical justification behind it. The idea behind the criteria is to select the feature $X_j \in S$ that is most redundant pair of $X_k$, so that if $X_k$ is redundant with any feature in $S$ it will have a low score. 


\subsection{High order estimators}
The category of high order approaches considers more than two order dependencies between the features. One of the pioneers is the Relax-mRMR method. Inspired by the mRMR score, the Relax-mRMR quantifies the interaction up to three features. The RelaxMRMR approximates the CMI by the following cost function
\begin{align}
 J^{\beta,\gamma,\eta}(X_k) &= I(X_k;Y) - \beta\sum_{X_j \in S_{}} I(X_j;X_k)+ \nonumber \gamma \sum_{X_j \in S_{}}I(X_j;X_k|Y) \nonumber \\&- \eta \sum_{X_j \in S_{k-1}}\sum_{X_i \in S_{k-1},i\neq j}I(X_k;X_i|X_j),
\end{align}
where different variants are presented for the weights $\beta, \gamma, \eta$ in the original work, where the averaging given by $J^{\text{RelaxMRMR}}(X_k)=J^{(\beta=|S|^{-1},\gamma=|S|^{-1},\eta=(|S||S-1|)^{-1})}(X_k)$ promotes the best results.

A method that takes more than three high order interactions {is the CMICOT} \cite{nips2016cmicot}, which is based on a min-max problem cost function \req{e.cmicot}:
\begin{align}
 J^{\text{CMICOT}}(X_k) = & \max_{H \subseteq S_{k-1}}\min_{G \subseteq S_{k-1}} I(X_k,H;Y|G),
 \label{e.cmicot} \\
 = & \max_{H \subseteq S_{k-1}} \left\{\rule[0.0em]{0.0em}{1.5em} I(X_k,H;Y) -\right. \label{e.cmicot.2} \left.\max_{G \subseteq S_{k-1}}\big[ I(X_k,H;G) - I(X_k,H;G|Y)\big]\right\},\nonumber
\end{align}
{where the objective is to find optimal sets $H, G\subseteq S_{k-1}$, where $H$ is referred to as an optimal complementary team of $X_k$, and $G\subseteq S_{k-1}$ is referred to as an optimal opposing team to $\{X_k\}\cup H$.} The method was inspired in the ensemble learning of decision trees, and the connection between $J^{\text{CMICOT}}(X_k)$ and $J^{\text{CMI}}(X_k)$ is not clear at first sight; however, the authors proved the convergence of $J^{\text{CMICOT}}(X_k) \rightarrow J^{\text{CMI}}(X_k)$ when $H$ and $G$ are the optimal solution of Eq.\ \req{e.cmicot}. 

More recently, \cite{Sechidis2019} have proposed high order versions of JMI and CMIM, considering $3^{rd}$ and $4^{rd}$ order interactions, which are stated as
\begin{align}
J^{\text{JMI-3}}(X_k) &= \sum_{X_j \in S} \sum_{ X_i \in S, i\neq j} I(X_j,X_i,X_k;Y), \\
J^{\text{JMI-4}}(X_k) &= \sum_{X_j \in S} \sum_{X_i \in S, i\neq j} \sum_{ X_t \in S, i\neq j, i\neq t, t\neq j} I(X_j,X_i,X_t,X_k;Y), \\
J^{\text{CMIM-3}}(X_k) &= \min_{X_j \in S, X_i \in S, i\neq j} I(X_k;Y|X_j,X_i).\\
J^{\text{CMIM-4}}(X_k) &= \min_{X_j, X_i, X_t \in S, i\neq j, i\neq t, t\neq j} I(X_k;Y|X_j,X_i,X_t).
\end{align}
The authors in \cite{Sechidis2019} derived these high order versions based on two main assumptions of pairwise conditional and class conditioned dependency among the set of selected and unselected features. Also, the JMI-3, JMI-4, CMIM-3, and CMIM-4 are computationally demanding due to the need to run over all possible combinations of $X_j$, $X_i$, and $X_t$. The JMI-3 and JMI-4 approximate the CMI by the sum of all possible combinations of the $3^{rd}$ and $4^{rd}$ orders. However, the relation of CMIM-3, and CMIM-4 scores to CMI is not straightforward to extract due to the minimum operator \cite{Sechidis2019}.

\section{High-Order approximation of CMI}
\label{sec.method}
The CMI estimation is not an easy task. It demands high order dependencies to be taken into account to correctly capture the feature dependencies. Under the umbrella of SFS for selecting $X_k$, the CMI is expanded as: 
\begin{align}
 I(X_k;Y|S)&= I(X_k;Y) - I(X_k;S) + I(X_k;S|Y),
\label{equ.J_CMIXXX} \\
&= I(X_k;Y) - R(X_k,S,Y).
\label{equ.costCMI2}
\end{align}
The last term, defined here as the total redundancy $R(X_k,S,Y)=I(X_k;S) - I(X_k;S|Y)$ contains all high order MI terms, while the first term (the relevance) measures the pairwise dependency $I(X_k;Y)$ and that is straightforward to estimate. To better understand the total redundancy, let rewriting the mutual information $I(X_k,S;Y)$ as
\begin{align}
I(X_k,S;Y)&=I(S;Y) + I(X_k;Y|S) \nonumber\\
&= I(S;Y)+I(X_k;Y) - I(X_k;S) + I(X_k;S|Y) \nonumber\\
&= \underbrace{I(S;Y)+I(X_k;Y)}_\text{sum of individual relevance}-\underbrace{R(X_k,S,Y)}_\text{total redundancy}.
\label{equ.TXkYZ}
\end{align}
A {positive $R(X_k,S,Y)$} decreases the mutual information $I(X_k,S;Y)$, as a result of the redundancy of $X_k$ being higher than the conditional redundancy. As consequence, a positive $R(X_k,S,Y)$ indicates that $X_k$ is redundant to $S$, and that the information of $X_k$, when combined with $S$, is less than the sum of the individual contributions of each variable $I(S;Y)+I(X_k;Y)$, implying that they have more information on $Y$ when separated than jointly. On the other {hand}, negative values of $R(X_k,S,Y)$ mean that $X_k$ jointly with $S$ adds up the {contribution of the individual information} of each variable w.r.t. $Y$, indicating that they have more information on $Y$ jointly that separated.

This led to the following statements.
\begin{mystat}
If $X_k$ is redundant with respect to the set $S$ \ie, if $X_k\in S$, then $I(X_k;Y|S)=0$ and $R(X_k,S,Y)=I(X_k;Y)$.
\end{mystat}
\begin{proof}
Since $X_k\in S$, then $X_k$ is constant conditioned on $S$, and therefore $I(X_k;Y|S)=0$ for any $Y$. Henceforth, from \req{equ.costCMI2} $R(X_k,S,Y)=I(X_k;Y)$.
\end{proof}

\begin{mystat}
If $X_k$ is independent from $S$, 
then $R(X_k,S,Y)=-I(X_k;S|Y)$, and $I(X_k;Y|S)=I(X_k;Y)+I(X_k;S|Y)$.
\end{mystat}
\begin{proof}
Since $X_k$ and $S$ are independent, then $I(X_k,S)=0$, $R(X_k,S,Y)=I(X_k;S)-I(X_k;S|Y)=-I(X_k;S|Y)$, and from \req{equ.J_CMI} $I(X_k;Y|S)=I(X_k;Y)+I(X_k;S|Y)$.
\end{proof}
Therefore, when a feature $X_k$ is independent of $S$ it only contributes to the $J_{}^{\text{CMI}}(\cdot)$ criterion when $X_k$ has (positive) conditional redundancy with respect to $S$. 

\begin{mystat}
The following relation holds $-I(X_k;S|Y)\leq R(X_k,S,Y)\leq I(X_k;Y)$.
\end{mystat}
\begin{proof}
This statement follows directly from the fact that the mutual information is non-negative.
\end{proof}
{From Statement 1, the CMI with a redundant feature will achieve the upper bound $I(X_k;Y)$ while, cf. Statement 2, an independent feature will reach the lower bound $-I(X_k;S|Y)$. The more redundant the feature $X_k$ is, the more positive $R(X_k,S,Y)$ becomes, up to the upper bound $I(X_k;Y)$.} 

The total redundancy term, which includes all high order dependencies, causes the most difficulty in estimating the CMI. The high order dependencies in $R(X_k,S,Y)$ are approximated here by setting a $n$th order representative subset $Z=\{Z _1,\ldots,Z _n\}$ of $S$, such that $Z\subseteq S $. In this case:
\begin{align}
{R}_n=R_n(X_k,Z,Y)=I(X_k,Z) - I(X_k,Z|Y)
\label{equ.Rn}
\end{align}\relax
where $R_n$ is the $n$th order approximation of the total redundancy.  
The elements in $Z$ are chosen to minimize the total redundancy difference between the complete and representative set, as follows:
\begin{align}
    Z &= \argmin_{{Z^* \subseteq S},\, |Z^*|=n} \Big[ {R(X_k,S,Y) - R_n(X_k,Z^*,Y) } \Big]\nonumber\\
      &= \argmin_{{Z^* \subseteq S},\, |Z^*|=n} { - R_n(X_k,Z^*,Y) } \nonumber\\
      &= \argmax_{{Z^* \subseteq S},\, |Z^*|=n} {  R_n(X_k,Z^*,Y) },
      \label{equ.maxR}
\end{align}
where the total redundancy $R(X_k,S,Y)$ was dropped from Eq. \req{equ.maxR} because is not a function of the representative set $Z$. By plugging the total redundancy with the representative set to the $n$th order approximation of CMI it becomes
\begin{align}
    {I}_n(X_k;Y|S) = I(X_k;Y) - {\max_{Z \subseteq S, |Z|=n} R^{}_n(X_k,Z,Y),}
    \label{equ.In}
\end{align}
where ${I}_n(X_k;S|Y)$ is the $n$th order approximation of CMI. 
Because the set $Z$ was chosen to maximize $R_n$, $Z$ can be interpreted as the most redundant subset of $S$ w.r.t. $X_k$ according to Statement 1. Instead, if $Z$ was chosen to minimize $R_n$, $Z$ would be the most complementary subset of $S$ with respect to $X_k$, according to Statement 2. This last possibility is also viable, as both directions converge to the total redundancy as $n \rightarrow |S|$, however, this last case is not wise from the feature selection point of view. 

The HOCMIM criterion is then defined as the $n$th order approximation of CMI:
\begin{align}
J^{\text{HOCMIM}}(X_k) &= {I}_n(X_k;Y|S)\nonumber\\
&=I(X_k;Y) - {\max_{Z \subseteq S, |Z|=n} R^{}_n(X_k,Z,Y),}\nonumber\\
&{=I(X_k;Y) - \max_{Z \subseteq S, |Z|=n} \left[  I(X_k;Z) - I(X_k;Z|Y)\right],}\nonumber\\
&{= I(X_k;Y) + \min_{Z \subseteq S, |Z|=n} \left[I(X_k;Z|Y) - I(X_k;Z)\right],}\nonumber\\
&=\min_{Z \subseteq S, |Z|=n} I(X_k;Y|Z)\label{equ.HOCMIM.1}
\end{align}
where $J^{\text{HOCMIM}}(X_k)\rightarrow J^{\text{CMI}}(X_k)$ as $n\rightarrow |S|$. Additionally, Eq. \req{equ.HOCMIM.1} enables the viewing of the application of MI for FS to high dimensions and also the equivalence with existing methods that employ high order dependencies. For instance, if $n=1$ the HOCMIM score Eq. \req{equ.HOCMIM.1} corresponds to the CMIM score \cite{Fleuret2004}. When $n=2$, and $n=3$ Eq. \req{equ.HOCMIM.1} corresponds to the CMIM-3 and CMIM-4 scores dicussed in \cite{Sechidis2019}, respectively; despite some notational ambiguity, CMIM-3 is equivalent to HOCMIM when $n=2$ and CMIM-4 is equivalent to HOCMIM when $n=3$.

For the HOCMIM, there is still the issue in estimating the representative set $Z$, and the order $n$. One possibility is to find the best subset $Z$ and $n$ by an exhaustive search of all possible solutions; however, this would incur a prohibitive computational cost. One solution is to keep the order $n$ constant and search exhaustively the best subset $Z$ of the order $n$; the exhaustive search over all possible solutions with size $n$ was used in \cite{Fleuret2004,Sechidis2019} for $n=1,2,3$, which remains computationally demand as $n$ grow, mainly for $n>2$.

The HOCMI overcomes existing limitations by sequentially selecting both the order $n$ and the best representative subset $Z$ using a greedy search. Let the total redundancy with the representative subset $Z$ be expanded into its chain form
\begin{align}
R_n(X_k,Z,Y) &= I(X_k;Z)-I(X_k;Z|Y), \label{equ.total.redundancy} \\
&=\sum_{i=1}^{|Z|} I(X_k;Z_i|Z_1,\ldots,Z_{i-1}) - \sum_{i=1}^{|Z|} I(X_k; Z_i|Z_1,\ldots,Z_{i-1}, Y)  . \nonumber 
\end{align}
with equality to $R(X_k,S,Y)$ iff $Z=S$. 
By setting the order $n$, the elements in $Z$ are found by a sequential greedy search. Specifically, the elements of $Z$
are found by the following greedy approximation, for $j=1,\ldots,n$:
\begin{align}
Z_j &= \argmax_{Z^*_j \in S} \left[ I(X_k;Z^*_j|{Z_1,\ldots,Z_{j-1}}) - I(X_k;Z^*_j|Y,{Z_1,\ldots,Z_{j-1}}) \right], \label{equ.Zmin}
\end{align}
and the order $n$ is selected so that
\begin{align}
n = \inf \left\{n^* \in \mathbb{Z}: I(X_k;Y) - R_{n^*}(X_k,Z,Y) < \epsilon \right\},
\label{equ.nsel}
\end{align}
where the difference $I(X_k;Y) - R_n(X_k,Z,Y)$ will always be positive because $I(X_k;Y)$ is the upper bound of total redundancy. In practice, for $I(X_k;Y) \neq 0$, the normalized distance $1 - {R_n(X_k,Z,Y)}/{I(X_k;Y)}$ is computed, and the threshold $0\leq \epsilon^* \leq 1$ must be defined, so that 
\begin{align}
n = \inf \left\{n^* \in \mathbb{P}: 1 - \frac{R_{n^*}(X_k,Z,Y)}{I(X_k;Y)} < \epsilon^* \right\}.
\label{equ.nsel2}
\end{align}
The Eq. \req{equ.nsel2} condition states that the elements in the representative set are chosen until they are sufficiently close to the upper bound. If feature $X_k$ is redundant, the algorithm will stop running the greedy search as soon as the feature's score approaches the upper bound, avoiding unnecessary computations. 
For the case of non-redundant features, the algorithms runs indefinitely (up to a upper limit), because there might have relevant high order dependencies that might contribute to CMI score.


Under these conditions, the next variable $X_k$ to be selected in the HOCMIM is then given by:
\begin{align}
    X_k &= \argmax_{X_j \in X\setminus S}\left[ \min_{Z \subseteq S, |Z|=n} I(X_k;Y|Z) \right],
\end{align}
where $n$ is selected as Eq. \req{equ.nsel}, and the elements $Z$ according to Eq. \req{equ.Zmin}. In practice, the solution of Eq. \req{equ.nsel2} also follows from a greedy search. The condition of Eq. \req{equ.nsel2}, $1 - \frac{R_{n^*}(X_k,Z,Y)}{I(X_k;Y)} < \epsilon^* $ is checked whenever a new element $Z_j$ is selected, and if satisfied, the algorithm stops selecting new elements for $Z$. When compared to other high order FS methods, this greedy search in the HOCMIM solution allows for a significant increase in the speed of feature selection, as well as selection of the order $n$ and elements $Z$ simultaneously. Along the experiments, the threshold $\epsilon^* =0.01$ was selected, also a upper limit to $n=15$ was considered.

Algorithm \ref{algo.hocmi} summarizes the HOCMIM FS method.
\begin{algorithm}[!t]
\footnotesize
\caption{HOCMIM}
\begin{algorithmic}[1]
\Procedure{HOCMIM}{$X$,$Y$,$K$}
\State{ $S_0 \leftarrow \emptyset$}
\Comment{Initialize set of selected features}
\State{ $X_k \leftarrow \argmax_{X^*_k \in S} I(X^*_k;Y)$}
\Comment{Select the index that maximizes MI}
\State{ $S_1 \leftarrow S_0 \cup X_{k}$}
\Comment{Add feature $X_{k}$ to the set of selected features}
\For{$i \gets 1$  to $K-1$}
\Comment{HOCMIM cost, Eq. \req{equ.HOCMIM.1}}
\State{ $X_k \leftarrow \argmax_{X^*_k \in X\setminus S_i} \min_{Z \subseteq S_i, |Z|=n} I(X^*_k;Y|Z)$}
\State{ $S_{i+1} \leftarrow S_i \cup X_{k}$}
\Comment{Add feature to the set of selected features}
\EndFor
\State{Return $S$}
\EndProcedure
\end{algorithmic}
\label{algo.hocmi}
\end{algorithm}\relax
The algorithm receives the input, target variable and the total number of features to select $K$. At each iteration, the greedy algorithm searches for the representative subset $Z$, together with checking the condition of Eq. \req{equ.nsel2}. As input to the algorithm, it receives the set of input variables, $X$, the target variable, $Y$, and the number of features to be selected $K$.  Finally, it returns as output the set of selected features $S$. A toy example of HOCMIM is discussed in Appendix A.

{\color{black} 


}

\subsection{MI estimation and Complexity}

The computation of MI is critical to estimate the high order dependencies. However, algorithms based on binning or maximum likelihood do not work well for high-order MI estimation. Many options can be found in the literature for MI estimation, for example the bias correction estimator \cite{Grassberger2003} or, for continuous variables, the k-nearest neighbor \cite{Kraskov2004}. Recently, efficient estimators for high-order MI have been proposed in \cite{Sechidis2019}. Due to the capabilities of the shrinkage estimators presented in \cite{Sechidis2019}, they will be used in our benchmarking experiments. 

{\color{black}
Assume the number of features to be selected is the top-$K$, and the total number of features is $D$. In \cite{Xuan2016}, the authors derived the complexity of low-order FS algorithms, such as mRMR, CMIM, JMI and DISR, which requires $\CMcal{O}(K^2D)$ calculations of MI. The third order algorithms RelaxMRMR, CMIM-3, JMI-3 requires the calculation of $\CMcal{O}(K^3D)$ MI terms, and the fourth order algorithms CMIM-4, JMI-4 requires to calculation of $\CMcal{O}(K^4D)$ MI terms. The CMICOT requires the which scales quadratically with the order $n$, requires the calculation of  $\CMcal{O}(n^2K^2D)$ MI components.

To select the top-$K$ features, the HOCMIM requires to the calculation of $\CMcal{O}(nK^2D)$ MI components. The derivation is as follows. The 1st iteration of HOCMIM, Algorithm \ref{algo.hocmi}, requires to calculate $D$ MI terms (the relevance terms). At the $k$th iteration, after selecting $k$ features, the HOCMIM requires to calculate $(D-k) + 2n(k-1)(D-k)$ MI components. The $(D-k)$ comes from the computation of the relevance $I(X_k;Y)$ for the $D-k$ remaining features. The next term $2n(k-1)(D-k)$ accounts for the cost of computing the total redundancy. The maximization stage in Eq. \req{equ.Zmin} requires $2n(k-1)$ MI calculations, for each of the $D-k$ remaining features, resulting in a total of $2n(k-1)(D-k)$ calculation of MI terms. To simplify, it is assumed that that $k$ is much smaller compared with $D$, $k\ll D$, then each iteration would require $D + 2n(k-1)D$ MI components. By summing over all $K$ iterations, it would result in a total of $KD + 2nD\frac{(K-1)K}{2}=KD(1 - n) + nK^2D$ MI calculations. By dropping the first element $KD(1 - n)$, which is smaller than the second for a large $K$, the number of MI components to be computed by the HOCMIM is then $\CMcal{O}(nK^2D)$.
}


In terms of {\color{black} MI calculations}, Table \ref{tab.complexity}
\begin{table}[!t]
\caption{\color{black} Complexity in terms of MI calculations for the different MI FS algorithms.}
\label{tab.datasets}
\footnotesize
\begin{center}
\begin{tabular}{lc}
\toprule
\multicolumn{2}{c}{\textbf{Complexity} }\\
\hline
mRMR, CMIM, JMI, DISR   &  $\CMcal{O}(K^2D)$\\
RelaxMRMR, CMIM-3, JMI-3 & $\CMcal{O}(K^3D)$ \\
CMIM-4, JMI-4 & $\CMcal{O}(K^4D)$ \\
CMICOT &   $\CMcal{O}(n^2K^2D)$\\
HOCMIM & $\CMcal{O}(nK^2D)$\\
\hline
\bottomrule
\end{tabular}
\end{center} 
\label{tab.complexity}
\end{table}\relax
summarizes the complexity of the state-of-art FS algorithms. The low order estimators mRMR, CMIM, JMI and DISR requires $\CMcal{O}(K^2d)$ MI computations, while the complexities of the high order estimators RelaxMRMR, CMIM-3, JMI-3, CMIM-4, JMI-4 scale exponentially with the order of dependencies. The CMICOT scales quadratically with the order of dependency $n$. On the other hand, the proposed HOCMIM scales linearly with $n$. This allows an efficient use of the HOCMIM in high-order FS approaches, being $n$ times more complex than standard low-order estimators.

\section{Experiments and Results}
\label{sec.experiments}

In this section, the HOCMIM is compared with the following state-of-the-art {\color{black}MI} FS methods CMIM \cite{Fleuret2004}, JMI \cite{Yang1999}, DISR \cite{Meyer2008}, mRMR \cite{Hanchuan2005}, Relax-mRMR \cite{Xuan2016}, CMICOT \cite{Shishkin2016}, {CMIM-3, CMIM-4, JMI-3 and JMI-4} \cite{Sechidis2019}, {\color{black} and the following state-of-the-art filter methods: Gini index, Fisher score, Kruskal-Wallis \cite{Kruska}, ILFS \cite{Roffo2017}, NCA \cite{NCA}, Relieff \cite{liu2008}, and the FS-OLS \cite{Zhang2022} }, in conjunction with the $k$-Nearest Neighbour (KNN) classifier, with neighbours set to $k=3$, and the Support Vector Machine with Linear Kernel (SVMlin) with the regularization parameter set to $C=1$. {\color{black} Using KNN with $k=3$ is a common practice in FS literature, since this classifier makes few assumptions about the data \cite{Brown2012}. The use of SVMlin with regularization parameter set to $C=1$ also follows previous FS research \cite{Nguyen2014,Xuan2016}. As the goal in this section is to quantify the quality of the FS algorithms rather than the performance of the classifiers, then it is assumed that the KNN and SVMlin with fixed parameters are sufficient  for evaluating the quality of subsets.} They are tested on a variety of classification benchmark datasets from {University of California, Irvine (UCI)} Machine Learning Repository \cite{Dua2019}.

%
A total of {\color{black}$20$} benchmark datasets were used in the experiments. The datasets are described in Table \ref{tab.datasets},
\begin{table}[!t]
\caption{Summary of the benchmark datasets.}
\label{tab.datasets}
{\tiny
\begin{center}
\setlength{\tabcolsep}{4 pt}
\begin{tabular}{l|c c c c}
\toprule
Dataset & n\# features ($D$)  & n\# classes ($C$) & n\# ex. ($N$) & ratio $N/D$ \\
\hline
parkinsonsEW & $22$ & $2$& $195$& $8.9$\\spectEW & $22$ & $2$& $267$& $12.1$\\german & $24$ & $2$& $1000$& $41.7$\\breastEW & $30$ & $2$& $569$& $19.0$\\wdbc & $31$ & $2$& $569$& $18.4$\\ionosphere & $33$ & $2$& $351$& $10.6$\\dermatology & $34$ & $6$& $358$& $10.5$\\soybeansmallEW & $35$ & $4$& $47$& $1.3$\\krvskpEW & $36$ & $2$& $3196$& $88.8$\\sonar & $60$ & $2$& $208$& $3.5$\\libras & $90$ & $15$& $360$& $4.0$\\semeionEW & $256$ & $10$& $1593$& $6.2$\\arrhythmia & $279$ & $13$& $452$& $1.6$\\ujiIndoor & $520$ & $3$& $21048$& $40.5$\\pcmac & $3289$ & $2$& $1943$& $0.6$\\basehock & $4862$ & $2$& $1993$& $0.4$\\gisette & $5000$ & $2$& $6000$& $1.2$\\pengleukEW & $7070$ & $2$& $72$& $0.0$\\gli-85 & $22283$ & $2$& $85$& $0.0$\\
\bottomrule
\end{tabular}
\end{center} 
}
\end{table}\relax
where the number ($n$\#) of features ($D$), classes ($C$), examples ($N$), and the ratio $N/D$ is indicated per dataset. These datasets vary in terms of number of examples, classes, features, and feature types. 

The HOCMIM and the Relax-mRMR algorithms are developed in Matlab\footnote{The source code of HOCMIM will be made available at the author’s github page \url{https://github.com/faasouza}}; the CMIM, JMI, DISR, and mRMR are based on the Feast Toolbox \cite{FeastToolbox}, written in C, while wrapper has been implemented in Matlab as well. The CMICOT is based on the author's implementation. The CMIM-3, CMIM-4, JMI-3, and JMI-4 were implemented from the author's source code. The MI and CMI computations in Relax-mRMR and HOCMIM were based on the estimator derived in \cite{Sechidis2019}.
{\color{black} The filter methods Gini index, Fisher score, Kruskal-Wallis test \cite{Kruska}, Relieff \cite{liu2008} are based on the Matlab implementation of the feature selection toolbox described in \cite{li2018feature}. The ILFS \cite{Roffo2017}, and FS-OLS \cite{Zhang2022} follows by the author's implementation. The NCA \cite{NCA} is based on the Matlab statistics and machine learning toolbox.
}

The continuous-valued features are discretized into $5$ bins by using an equal-width strategy. The data is split randomly into training-testing sets ($50$\% for training and the remaining $50$\% for testing); this process is repeated for 30 times. The first $50$ features are chosen from the training data, and their performance is assessed with the KNN and the SVMlin classifiers using the testing data. Then, the cross-validation error vs.\ the number of selected features ($|S|$) are computed and the average of cross-validation misclassification error for all $50$ features  (over all 30 trials) are taken and used for comparison .

In the experiments, the Nemenyi post-hoc test is employed to compare the multiple FS algorithms over the benchmark datasets; the Friedman test is used to access the global significance. The critical diagram (CD) \cite{Demsar20016} is used as a visualization aid to check the statistical difference of the FS algorithms over the Nemenyi test. The interpretation of CD is straightforward (check Fig. \ref{fig.cd_knn_all_methods} for a view on a CD diagram), the top line axis indicates the average rank of the FS methods over all datasets; for each dataset, the best (lowest misclassification error) FS algorithm gets the rank of $1$, the second-best rank $2$, and so on, in case of ties average ranks are assigned. The ranks are sorted in such a way that the lowest (best) ranks are to the right, on the other hand, the methods with the highest ranks (worse) are to the left. The FS algorithms’ groups connected by a line are statistically similar at the level $\alpha=0.1$. On the top of CD diagram, there is an interval that indicates the critical distance. The FS algorithms separated by a distance greater than the critical distance have a statistically significant difference in performance.

\subsection{Performance on {\color{black} low dimensional} datasets}
\label{subsec_bench_results}
{\color{black}In this section, the results of the FS methods on the low dimensional ($D \leq 520$) benchmark datasets are discussed. }

The CD for all {\color{black}the MI} FS methods is shown in Fig.\ \ref{fig.cd_knn_all_methods},
\begin{figure*}[!t]
\centering
\subfigure[]
{\label{fig.cd_knn_all_methods}\includegraphics[width=.49\columnwidth]{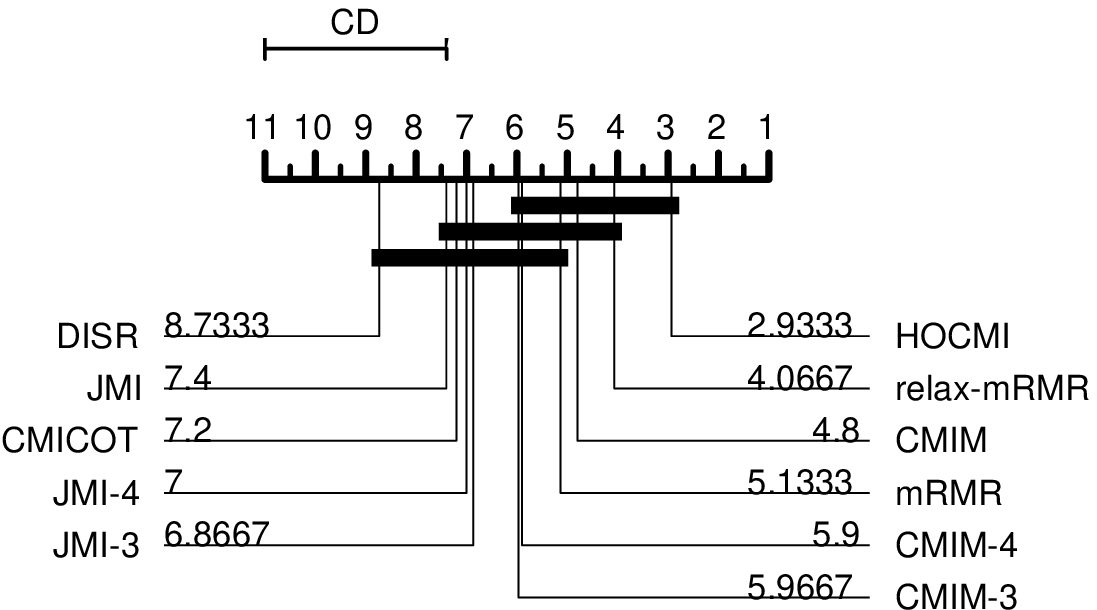}}
\subfigure[]
{\label{fig.cd_svmlin_all_methods}\includegraphics[width=.485\columnwidth]{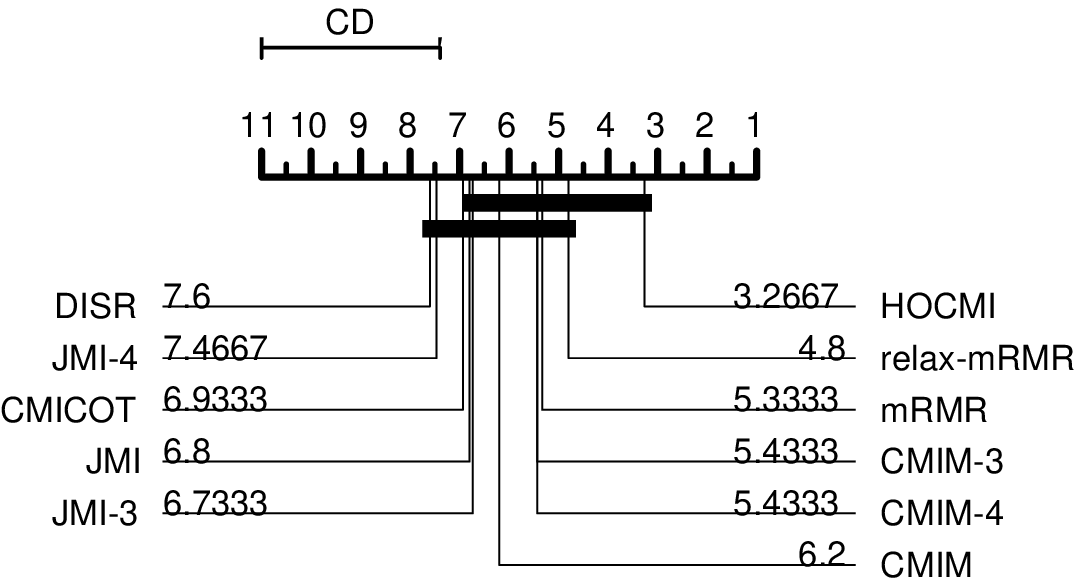}}
\caption{CD on the UCI benchmarking datasets, using the a) KNN classifier, b) SVM-lin classifier.}
\label{fig.all_methods}
\end{figure*}\relax
for the KNN classifier, and Fig.\ \ref{fig.cd_svmlin_all_methods} for the SVMlin classifier. In complement to the CD diagram, Table \ref{tab.results}
\begin{table*}[!t]
\caption{Average misclassification error over $30$ repetitions and the rank of the FS methods for the KNN classifier.}
\centering
\label{tab.results}
\tiny
\setlength{\tabcolsep}{0.7 pt}
\begin{tabular}{|l|c|c|c|c|c|c|c|c|c|c|c|}
\toprule
& \rot{\textbf{HOCMIM}} & \rot{\textbf{CMIM}} & \rot{\textbf{JMI}} & \rot{\textbf{DISR}} & \rot{\textbf{mRMR}} & \rot{\textbf{CMIM-3}} & \rot{\textbf{CMIM-4}}  & \rot{\textbf{JMI-3}}  & \rot{\textbf{JMI-4}}  & \rot{\textbf{relax-mRMR}}  & \rot{\textbf{CMICOT}}   \\ 
\hline
parkinsonsEW & $0.138(1)$ & $0.138(2)$ & $0.150(10)$ & $0.158(11)$ & $0.146(8)$ & $0.141(4)$ & $0.142(6)$ & $0.145(7)$ & $0.148(9)$ & $0.142(5)$  & $0.139(3)$\\spectEW & $0.215(3)$ & $0.218(5)$ & $0.223(10)$ & $0.226(11)$ & $0.215(2)$ & $0.219(6)$ & $0.216(4)$ & $0.220(8)$ & $0.220(9)$ & $0.214(1)$  & $0.219(7)$\\german & $0.300(3)$ & $0.301(6)$ & $0.300(4)$ & $0.303(11)$ & $0.301(8)$ & $0.296(1)$ & $0.298(2)$ & $0.301(7)$ & $0.301(5)$ & $0.302(10)$  & $0.301(9)$\\breastEW & $0.068(4)$ & $0.067(2)$ & $0.071(6)$ & $0.082(11)$ & $0.065(1)$ & $0.069(5)$ & $0.072(7)$ & $0.074(9)$ & $0.077(10)$ & $0.067(3)$  & $0.074(8)$\\wdbc & $0.056(4)$ & $0.056(3)$ & $0.061(9)$ & $0.076(11)$ & $0.054(1)$ & $0.057(5)$ & $0.057(6)$ & $0.060(8)$ & $0.062(10)$ & $0.055(2)$  & $0.058(7)$\\ionosphere & $0.150(4)$ & $0.151(5)$ & $0.152(7)$ & $0.143(1)$ & $0.151(6)$ & $0.155(10)$ & $0.147(2)$ & $0.155(9)$ & $0.154(8)$ & $0.149(3)$  & $0.161(11)$\\dermatology & $0.105(1)$ & $0.109(3)$ & $0.145(10)$ & $0.146(11)$ & $0.141(9)$ & $0.106(2)$ & $0.111(4)$ & $0.128(8)$ & $0.116(6)$ & $0.118(7)$  & $0.115(5)$\\soybeansmallEW & $0.042(7)$ & $0.049(8)$ & $0.031(5)$ & $0.031(3)$ & $0.031(2)$ & $0.073(9)$ & $0.073(10)$ & $0.031(1)$ & $0.031(4)$ & $0.031(6)$  & $0.086(11)$\\krvskpEW & $0.071(1)$ & $0.092(9)$ & $0.087(5)$ & $0.092(8)$ & $0.090(7)$ & $0.100(11)$ & $0.098(10)$ & $0.082(4)$ & $0.079(2)$ & $0.088(6)$  & $0.081(3)$\\sonar & $0.221(1)$ & $0.226(3)$ & $0.230(5)$ & $0.243(10)$ & $0.226(4)$ & $0.231(7)$ & $0.231(6)$ & $0.235(8)$ & $0.243(11)$ & $0.224(2)$  & $0.240(9)$\\libras & $0.402(2)$ & $0.403(3)$ & $0.507(10)$ & $0.524(11)$ & $0.420(5)$ & $0.425(6)$ & $0.445(7)$ & $0.457(9)$ & $0.447(8)$ & $0.397(1)$  & $0.418(4)$\\semeionEW & $0.308(4)$ & $0.304(3)$ & $0.413(10)$ & $0.429(11)$ & $0.396(9)$ & $0.294(2)$ & $0.291(1)$ & $0.390(8)$ & $0.371(7)$ & $0.370(6)$  & $0.319(5)$\\arrhythmia & $0.395(7)$ & $0.399(8)$ & $0.383(6)$ & $0.373(1)$ & $0.381(4)$ & $0.412(9)$ & $0.423(10)$ & $0.382(5)$ & $0.381(3)$ & $0.380(2)$  & $0.425(11)$\\penglungEW & $0.327(1)$ & $0.330(2)$ & $0.350(6)$ & $0.356(9)$ & $0.345(5)$ & $0.333(3)$ & $0.351(7)$ & $0.354(8)$ & $0.362(11)$ & $0.339(4)$  & $0.362(10)$\\ujiIndoor & $0.052(1)$ & $0.073(10)$ & $0.066(8)$ & $0.074(11)$ & $0.061(6)$ & $0.067(9)$ & $0.062(7)$ & $0.058(4)$ & $0.055(2)$ & $0.055(3)$  & $0.058(5)$\\
 \hline
Average rank & $2.88$ & $4.69$ & $7.56$ & $8.88$ & $5.12$ & $5.94$ & $6.00$ & $7.00$ & $7.06$ & $3.88$& $7.00$  \\
\bottomrule
\end{tabular}
\end{table*}
shows the average misclassification error obtained on the benchmark datasets for the KNN classifier, and Table \ref{tab.resultssvm}
\begin{table*}[!t]
\caption{Average misclassification error over $30$ repetitions and the rank of the FS methods for the SVM-lin classifier.}
\centering
\label{tab.resultssvm}
\tiny
\setlength{\tabcolsep}{0.7 pt}
\begin{tabular}{|l|c|c|c|c|c|c|c|c|c|c|c|}
\toprule
& \rot{\textbf{HOCMIM}} & \rot{\textbf{CMIM}} & \rot{\textbf{JMI}} & \rot{\textbf{DISR}} & \rot{\textbf{mRMR}} & \rot{\textbf{CMIM-3}} & \rot{\textbf{CMIM-4}}  & \rot{\textbf{JMI-3}}  & \rot{\textbf{JMI-4}}  & \rot{\textbf{relax-mRMR}}  & \rot{\textbf{CMICOT}}   \\ 
\hline
parkinsonsEW & $0.337(5)$ & $0.338(8)$ & $0.337(2)$ & $0.337(4)$ & $0.335(1)$ & $0.338(6)$ & $0.338(7)$ & $0.341(10)$ & $0.341(11)$ & $0.341(9)$  & $0.337(3)$\\spectEW & $0.377(1)$ & $0.382(10)$ & $0.378(3)$ & $0.378(5)$ & $0.379(7)$ & $0.381(9)$ & $0.378(6)$ & $0.377(2)$ & $0.378(4)$ & $0.380(8)$  & $0.383(11)$\\german & $0.325(5)$ & $0.326(7)$ & $0.325(3)$ & $0.329(11)$ & $0.326(8)$ & $0.324(1)$ & $0.325(2)$ & $0.325(4)$ & $0.325(6)$ & $0.327(10)$  & $0.326(9)$\\breastEW & $0.054(3)$ & $0.055(4)$ & $0.059(8)$ & $0.066(11)$ & $0.054(2)$ & $0.056(5)$ & $0.058(7)$ & $0.062(9)$ & $0.064(10)$ & $0.053(1)$  & $0.057(6)$\\wdbc & $0.043(6)$ & $0.042(3)$ & $0.046(9)$ & $0.055(11)$ & $0.040(1)$ & $0.043(4)$ & $0.043(5)$ & $0.046(8)$ & $0.047(10)$ & $0.042(2)$  & $0.045(7)$\\ionosphere & $0.180(4)$ & $0.202(7)$ & $0.235(9)$ & $0.155(1)$ & $0.156(2)$ & $0.217(8)$ & $0.192(5)$ & $0.252(11)$ & $0.237(10)$ & $0.159(3)$  & $0.200(6)$\\dermatology & $0.095(3)$ & $0.101(4)$ & $0.138(10)$ & $0.142(11)$ & $0.133(9)$ & $0.094(1)$ & $0.095(2)$ & $0.127(8)$ & $0.120(7)$ & $0.115(6)$  & $0.102(5)$\\soybeansmallEW & $0.038(6)$ & $0.047(10)$ & $0.036(4)$ & $0.035(2)$ & $0.036(3)$ & $0.045(8)$ & $0.045(9)$ & $0.037(5)$ & $0.039(7)$ & $0.034(1)$  & $0.053(11)$\\krvskpEW & $0.066(1)$ & $0.068(10)$ & $0.068(8)$ & $0.067(5)$ & $0.068(7)$ & $0.068(9)$ & $0.069(11)$ & $0.067(4)$ & $0.066(3)$ & $0.067(6)$  & $0.066(2)$\\sonar & $0.298(6)$ & $0.300(7)$ & $0.291(1)$ & $0.302(9)$ & $0.301(8)$ & $0.297(5)$ & $0.293(2)$ & $0.294(3)$ & $0.303(10)$ & $0.296(4)$  & $0.306(11)$\\libras & $0.545(3)$ & $0.543(2)$ & $0.619(10)$ & $0.631(11)$ & $0.558(5)$ & $0.561(6)$ & $0.582(9)$ & $0.580(8)$ & $0.573(7)$ & $0.541(1)$  & $0.555(4)$\\semeionEW & $0.313(3)$ & $0.323(4)$ & $0.420(10)$ & $0.437(11)$ & $0.404(9)$ & $0.308(2)$ & $0.301(1)$ & $0.399(8)$ & $0.381(7)$ & $0.381(6)$  & $0.336(5)$\\arrhythmia & $0.409(1)$ & $0.450(8)$ & $0.438(7)$ & $0.417(2)$ & $0.425(3)$ & $0.464(10)$ & $0.459(9)$ & $0.430(6)$ & $0.427(5)$ & $0.426(4)$  & $0.484(11)$\\penglungEW & $0.423(1)$ & $0.426(2)$ & $0.455(8)$ & $0.455(9)$ & $0.440(6)$ & $0.428(3)$ & $0.430(4)$ & $0.449(7)$ & $0.461(10)$ & $0.433(5)$  & $0.473(11)$\\ujiIndoor & $0.048(1)$ & $0.066(7)$ & $0.075(10)$ & $0.082(11)$ & $0.070(9)$ & $0.058(4)$ & $0.055(3)$ & $0.068(8)$ & $0.065(5)$ & $0.065(6)$  & $0.053(2)$\\
 \hline
Average rank & $3.27$ & $6.20$ & $6.80$ & $7.60$ & $5.33$ & $5.40$ & $5.47$ & $6.73$ & $7.47$ & $4.80$& $6.93$  \\
\bottomrule
\end{tabular}
\end{table*}
for the SVMlin classifier.
At the side of the average misclassification error, the number inside the parenthesis indicates the rank of each algorithm with respect to the misclassification error; the best FS algorithm gets the rank of $1$, the second best rank $2$, and so on. At the bottom of the tables the average rank error {\color{black} from the Friedman test} is indicated, allowing a quick assessment of results. {\color{black} The statistical significance of the average rank should be accessed together the critical diagram. 
Figure \ref{fig.error_lowdim_plot} 
\begin{figure*}[!t]
\centering
\subfigure[]
{\label{fig.knn_error_plot_average_all_lowdim}\includegraphics[width=0.49\columnwidth]{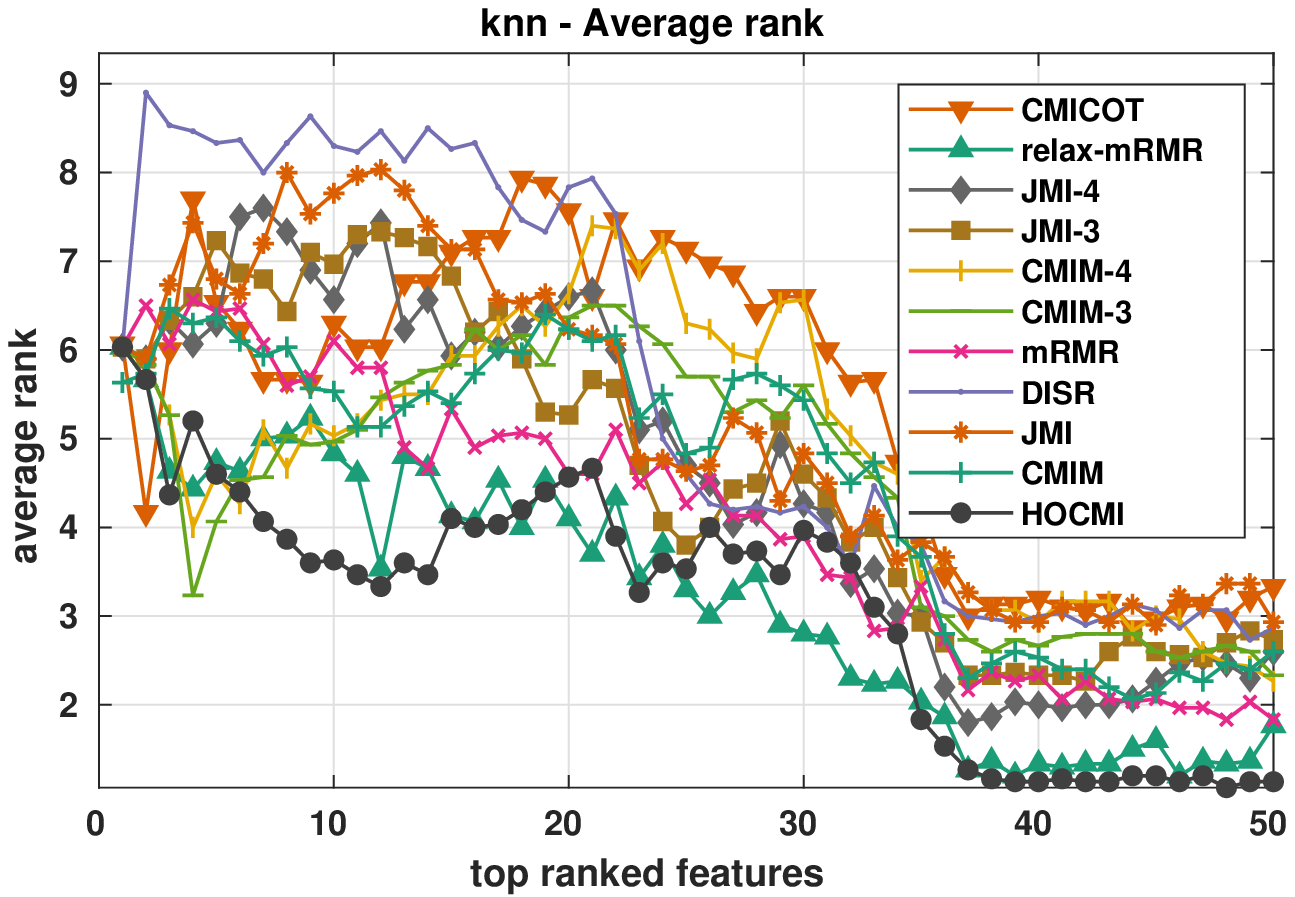}}
\subfigure[]
{\label{fig.svmlin_error_plot_average_all_lowdim}\includegraphics[width=0.49\columnwidth]{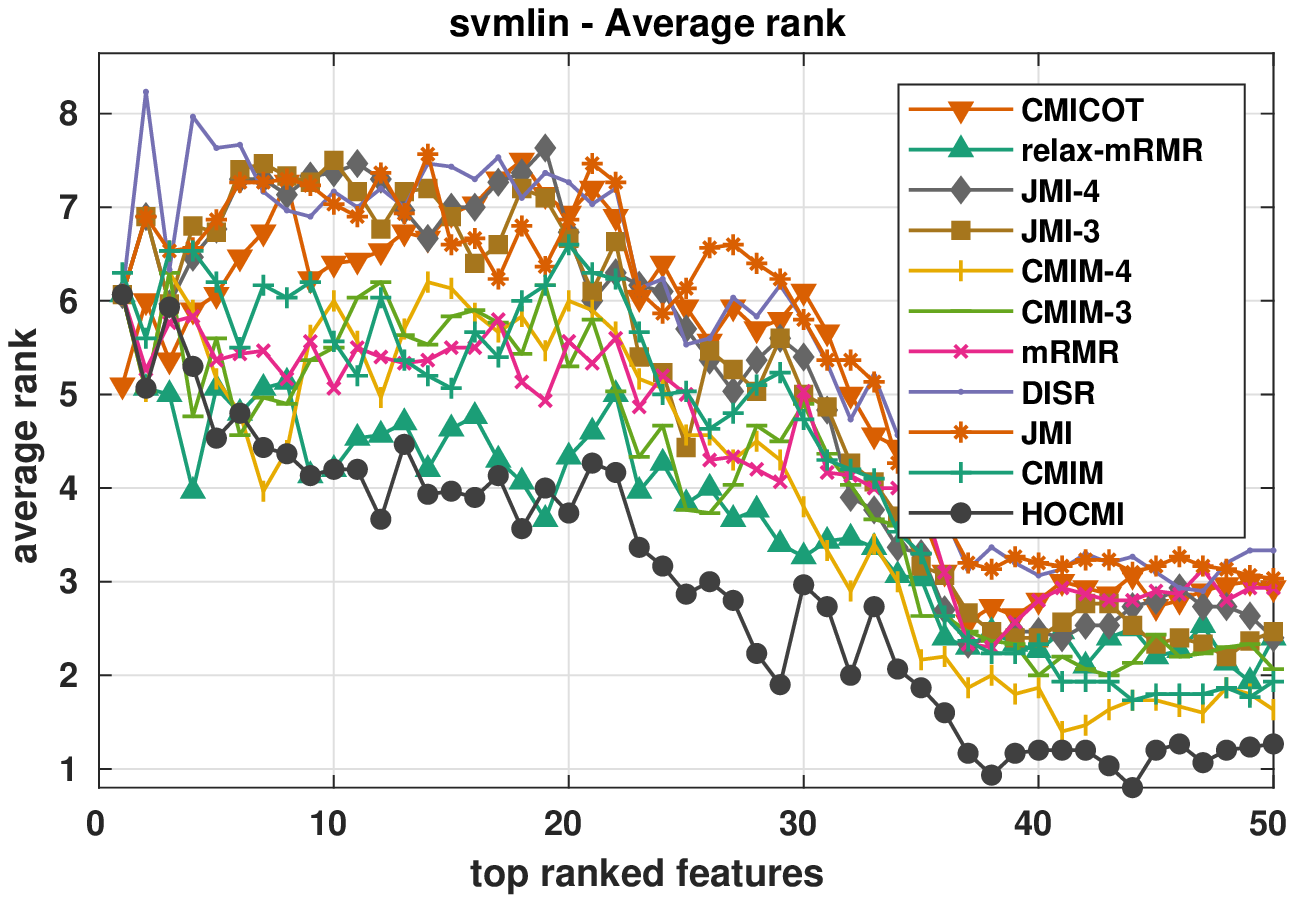}} 
\caption{Top ranked features vs the rank of the MI FS algorithms (averaged over all low dimensional datasets) for a) KNN and b) SVMlin classifiers.}
\label{fig.error_lowdim_plot}
\end{figure*}\relax
shows the line charts with the average ranking (over all low dimensional datasets) along side with the top-ranked features for all feature selection methods, for the KNN (Fig. \ref{fig.knn_error_plot_average_all_lowdim}) and SVMlin (Fig. \ref{fig.svmlin_error_plot_average_all_lowdim}) classifiers. 
}

First, by checking the results of the CD diagram from Figs.\ \ref{fig.cd_knn_all_methods} and \ref{fig.cd_svmlin_all_methods} shows the superiority of HOCMIM over all other FS methods, it has an average rank of $2.88$ and $3.27$ for the KNN and SVMlin classifiers, followed by the relax-mRMR with an average rank of $4.06$, and $4.80$ for the KNN and SVMlin classifiers, respectively. The low order estimators CMIM and mRMR, and the high order methods CMIM-3 and CMIM-4 have scores between $4.80$ and $6.20$, and interchangeable ranks for both classifiers. Also, according to the CD diagram, the HOCMIM is statistically similar in either the classifiers, compared to the relax-mRMR, CMIM-3, CMIM-3, mRMR, and CMIM; despite still providing the lowest rank score for both classifiers.
For the KNN classifier, the HOCMIM is statistically superior to the JMI-3, JMI-4, CMIMCOT, JMI, and DISR methods. On the other hand, for the SVMlin classifier, the HOCMIM is statistically superior to the CMICOT, JMI-4, and DISR methods.



{\color{black} 
Examining the rank plot over the top ranked features in  Fig.\ \ref{fig.svmlin_error_plot_average_all_lowdim} shows that the HOCMIM ranks lower than the other methods for the majority of the top-ranked features for the KNN and SVMlin classifiers. This is more noticeable in the final top-ranked features. The relax-mRMR is the most comparable (standing in second) to the HOCMIM, this behaviour is more evident for the last top ranked features.} From the results in Table \ref{tab.results}, {\color{black}and Table \ref{tab.resultssvm}}, it is possible to observe that the HOCMIM, and high order methods has the tendency to perform better in the datasets with higher $N/D$ ratio, take for instance the ``ujiIndoor'', and ``krvskpEW'' datasets, where the HOCMIM, CMIM-3, CMIM-4, JMI-4, JMI-4 and relax-mRMR stands among the top rank positions. Similarly, the high order methods (except the HOCMIM), stands in the last top ranked positions for datasets with low $N/D$ ratio, see for instance the ``penglungEW'' and ``soybeansmallEW'' datasets. 
{\color{black}
From Tables \ref{tab.results} and \ref{tab.resultssvm} it is possible to observe some discrepancies for some datasets, for example, the ``parkinsonEW'' where the HOCMIM stands as rank 1 for the KNN, and the SVM-lin stands as rank 5. Also the ``arrhythmia'' dataset, where the HOCMIM ranks as $7$ for the KNN, and $1$ for the SVM-lin. The main difference between both datasets is the total number of classes, while the ``parkinsonEW'' has $2$ classes, the ``arrhythmia'' contains $13$ classes. It is possible to see that when the dataset has a small ratio $N/D$, the HOCMIM subset trained with SVMlin outperforms the KNN. This is most likely due to SVM's optimality for small-sample classification problems.

}

Furthermore, despite the overall good performance of the HOCMIM, it also stands because of the low computational cost compared the order high order methods. 

\subsection{\color{black} Performance on high dimensional datasets}
{\color{black}
Five additional experiments were carried out in order to assess the performance of the HOCMIM in high-dimensional datasets. The experiments used the datasets "pcmac", "basehock", "gisette", "pengleuk", and "gli-85". For comparison, the low order FS algorithms were run, together with the HOCMIM and CMICOT, whose complexity grows linearly, and quadratically with the order of dependencies, respectively.



The results for the KNN classifiers are indicated in Table \ref{tab.highdim_knn}, {\color{black}
\begin{table*}[!t]
\caption{Average misclassification error over $30$ repetitions and the rank of the FS methods for the KNN classifier in the high dimensional datasets.}
\centering
\label{tab.highdim_knn}
\tiny
\setlength{\tabcolsep}{0.7 pt}
\begin{tabular}{|l|c|c|c|c|c|c|}
\toprule
& \rot{\textbf{HOCMIM}} & \rot{\textbf{CMIM}} & \rot{\textbf{JMI}} & \rot{\textbf{DISR}} & \rot{\textbf{mRMR}} & \rot{\textbf{CMICOT}} \\ 
\hline
pcmac& $0.333(1)$& $0.334(3)$& $0.350(5)$& $0.338(4)$& $0.334(2)$& $0.364(6)$\\basehock& $0.316(1)$& $0.322(3)$& $0.336(4)$& $0.346(5)$& $0.321(2)$& $0.349(6)$\\gisette& $0.068(2)$& $0.077(4)$& $0.076(3)$& $0.081(6)$& $0.080(5)$& $0.064(1)$\\pengleukEW& $0.056(3)$& $0.056(4)$& $0.047(1)$& $0.059(5)$& $0.049(2)$& $0.191(6)$\\gli-85& $0.175(3)$& $0.174(2)$& $0.199(5)$& $0.182(4)$& $0.165(1)$& $0.284(6)$\\
 \hline
Average & $2.0$& $3.2$& $3.6$& $4.8$& $2.4$& $5.0$\\
\bottomrule
\end{tabular}
\end{table*}
and for the SVMlin classifier are indicated in Table \ref{tab.highdim_svmlin}.
\begin{table*}[!ht]
\caption{Average misclassification error over $30$ repetitions and the rank of the FS methods for the SVM-lin classifier in the high dimensional datasets.}
\centering
\label{tab.highdim_svmlin}
\tiny
\setlength{\tabcolsep}{0.7 pt}
\begin{tabular}{|l|c|c|c|c|c|c|}
\toprule
& \rot{\textbf{HOCMIM}} & \rot{\textbf{CMIM}} & \rot{\textbf{JMI}} & \rot{\textbf{DISR}} & \rot{\textbf{mRMR}} & \rot{\textbf{CMICOT}} \\ 
\hline
pcmac& $0.333(1)$& $0.334(3)$& $0.350(5)$& $0.338(4)$& $0.334(2)$& $0.364(6)$\\basehock& $0.310(3)$& $0.313(4)$& $0.300(2)$& $0.298(1)$& $0.316(5)$& $0.328(6)$\\gisette& $0.110(4)$& $0.100(1)$& $0.115(6)$& $0.107(3)$& $0.103(2)$& $0.112(5)$\\pengleukEW& $0.056(3)$& $0.056(4)$& $0.047(1)$& $0.059(5)$& $0.049(2)$& $0.191(6)$\\gli-85& $0.237(1)$& $0.247(2)$& $0.262(5)$& $0.261(4)$& $0.254(3)$& $0.323(6)$\\
 \hline
Average & $2.4$& $2.8$& $3.8$& $3.4$& $2.8$& $5.8$\\
\bottomrule
\end{tabular}
\end{table*}\relax
On average rank, HOCMIM outperforms all other methods, according to both tables, standing with average rank of $2.0$ for the KNN and $2.4$ for the SVMlin classifier. To complement the analysis, the critical diagram, for both classifiers are shown in Fig. \ref{fig.cd_highdim}.
\begin{figure*}[!t]
\centering
\subfigure[]
{\label{fig.cd_knn_highdim}\includegraphics[width=.48\columnwidth]{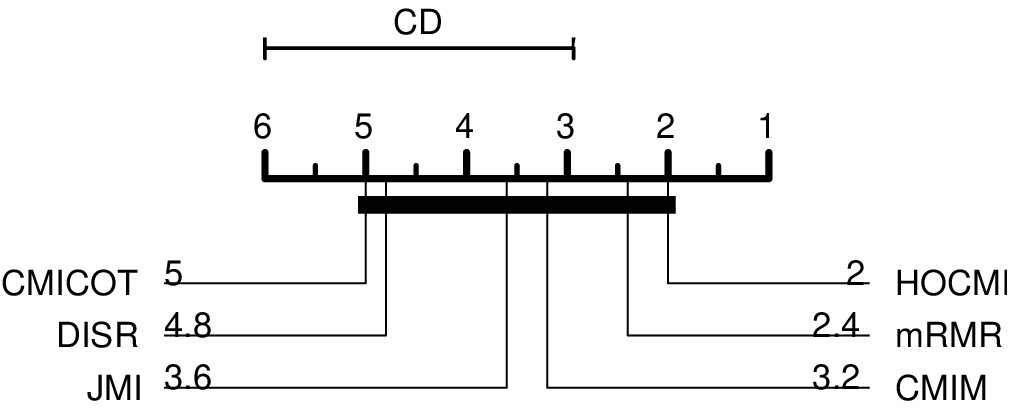}}
\subfigure[]
{\label{fig.cd_svmlin_highdim}\includegraphics[width=.48\columnwidth]{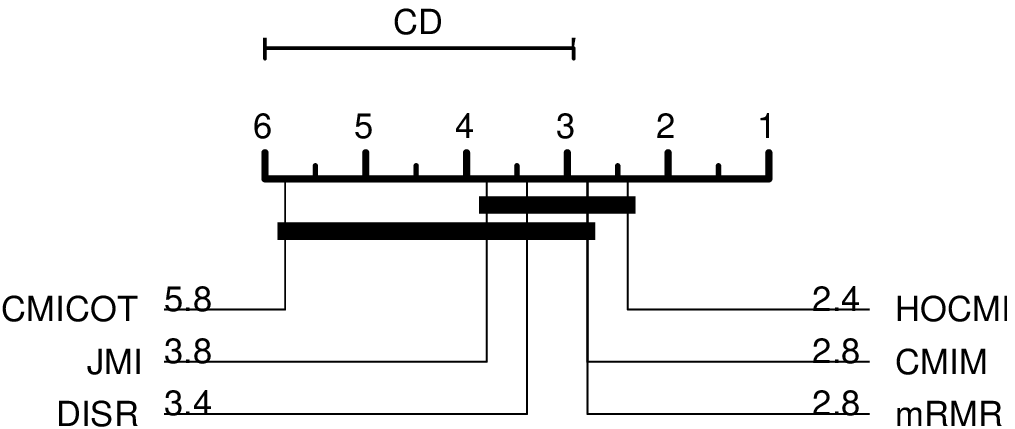}}
\caption{CD on the high dimensional datasets, using the a) KNN classifier, b) SVM-lin classifier.}
\label{fig.cd_highdim}
\end{figure*}\relax
The CD results of the KNN classifiers show that all methods are statistically similar to each other. All methods are statistically similar for the SVM-lin classifier, with the exception of the CMICOT, in which the HOCMIM statistically superior.

For the KNN classifier, Table \ref{tab.highdim_knn} the HOCMIM has the tendency to perform better when the dataset has a large number of samples ($N>1000$). Take for instance the ``pcmac'', and ``basehock'' and ``gisette'' datasets. On the other hand, for the $D\ll N$ datasets, the low dimensional methods has the tendency to work better. This also has to do with the effect of the MI estimator, which can still be prone to variance and bias, in which can affect the quality of the final selected features. For the SVM classifier (Table \ref{tab.highdim_svmlin}) this pattern is not observed, as the ``pcmac'', and ``gli-85'' datasets perform better. This behavior is the similar to the one discussed in previous section, and has to do with optimally of SVM for small-sample classifications applications.
The line charts indicating the average rank over the five datasets are indicated in Fig. \ref{fig.error_plot_average_all_highdim}
\begin{figure*}[!t]
\centering
\subfigure[]
{\label{fig.knn_error_plot_average_all_highdim}\includegraphics[width=0.49\columnwidth]{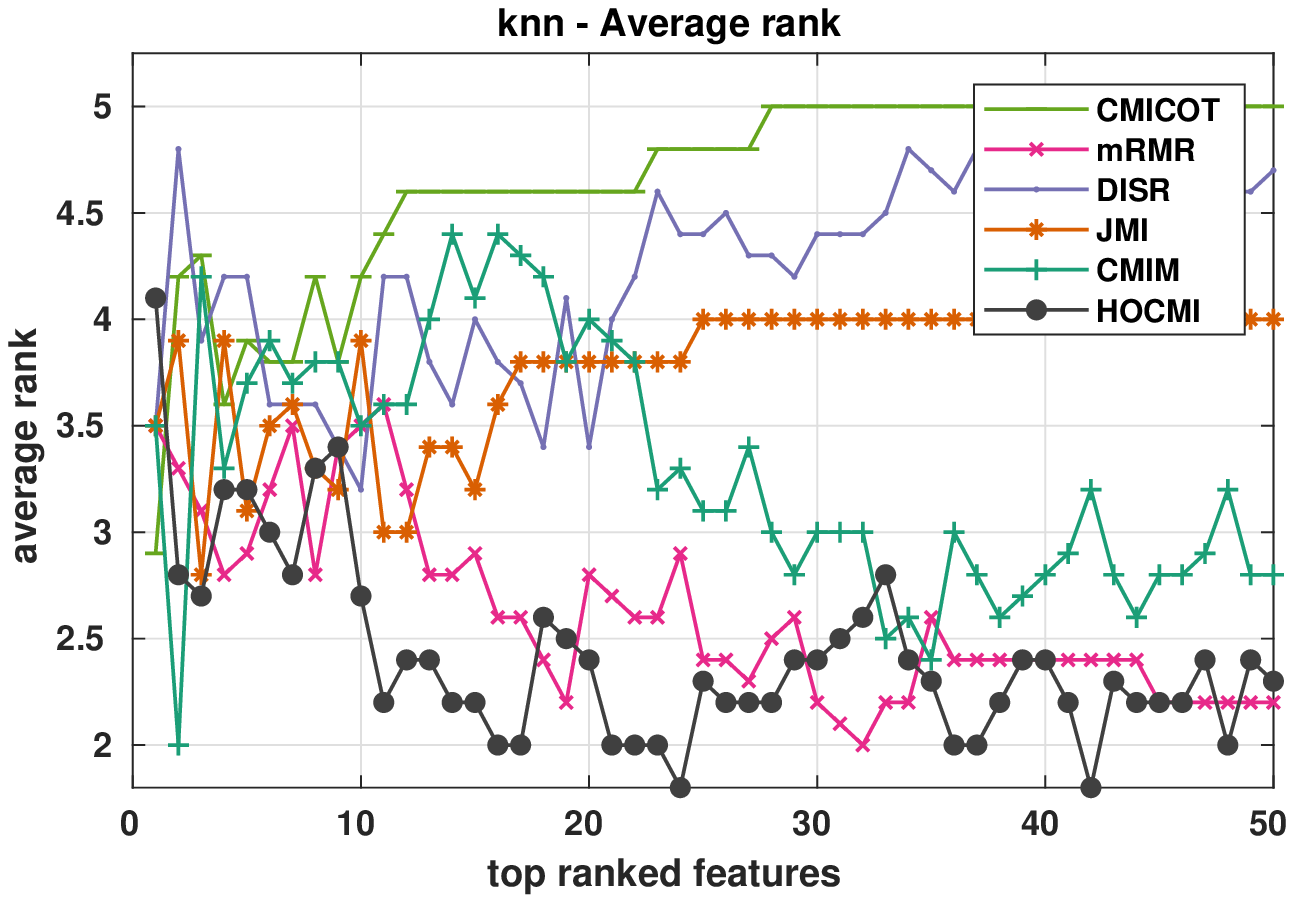}}
\subfigure[]
{\label{fig.svmlin_error_plot_average_all_highdim}\includegraphics[width=0.49\columnwidth]{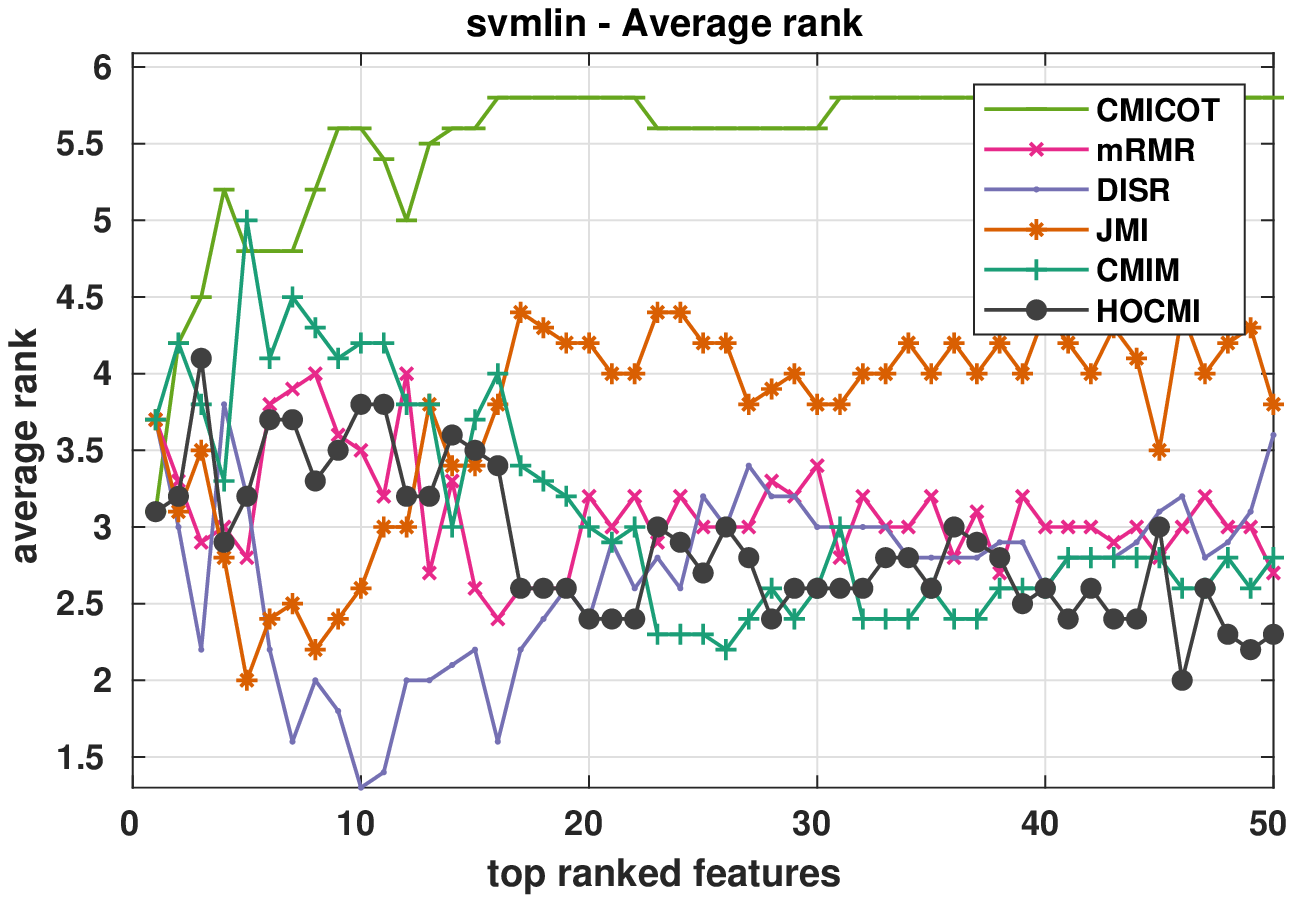}} 
\caption{Top ranked features vs the rank of the MI FS algorithms (averaged over all high dimensional) for a) KNN and b) SVMlin classifiers.}
\label{fig.error_plot_average_all_highdim}
\end{figure*}\relax
For the KNN classifier, the average rank plot (Fig. \ref{fig.knn_error_plot_average_all_highdim}) of HOCMIM is the lowest in the majority of the top ranked features, with the mRMR standing in second. In the case of the SVM-lin classifier (Fig. \ref{fig.svmlin_error_plot_average_all_highdim}), the average rank is more balanced, with the HOCMIM performing poorly in the first top ranked features, and the DISR scoring the lowest ranking. From the top-ranked features $20$ and up, the HOCMIM and relax-mRMR perform better  than the DISR.

}

}

\subsection{\color{black} Comparison with other filter methods}
{\color{black} 
In this section, the performance of the HOCMIM is compared with to the state of the art filter feature selection methods, namely the Relieff, ILFS, Fischer, NCA, FS-OLS, and Gini. For these methods, all datasets from Table \ref{tab.datasets} were evaluated. 
 
The misclassification error for for the HOCMIM and the other filter FS methods are exhibited in Table \ref{tab.knn_results_allfilter}
\begin{table*}[!t]
\caption{Average misclassification error over $30$ repetitions and the rank of the FS methods for the KNN classifier.}
\centering
\label{tab.knn_results_allfilter}
\tiny
\setlength{\tabcolsep}{0.7 pt}
\begin{tabular}{|l|c|c|c|c|c|c|c|c|}
\toprule
& \rot{\textbf{HOCMIM}} & \rot{\textbf{RELIEFF}} & \rot{\textbf{ILFS}} & \rot{\textbf{Fischer}} & \rot{\textbf{NCA}} & \rot{\textbf{FS-OLS}}  & \rot{\textbf{Kruska}}  & \rot{\textbf{GINI}}   \\ 
\hline
parkinsonsEW& $0.138(1)$& $0.141(4)$& $0.139(2)$& $0.148(6)$& $0.139(3)$& $0.142(5)$& $0.169(8)$& $0.150(7)$\\spectEW& $0.215(1)$& $0.218(4)$& $0.217(3)$& $0.225(7)$& $0.218(5)$& $0.217(2)$& $0.225(6)$& $0.225(8)$\\german& $0.300(2)$& $0.304(5)$& $0.318(7)$& $0.301(4)$& $0.305(6)$& $0.298(1)$& $0.327(8)$& $0.301(3)$\\breastEW& $0.068(1)$& $0.078(6)$& $0.092(7)$& $0.072(3)$& $0.075(4)$& $0.069(2)$& $0.093(8)$& $0.077(5)$\\wdbc& $0.056(2)$& $0.061(3)$& $0.082(7)$& $0.064(5)$& $0.066(6)$& $0.056(1)$& $0.217(8)$& $0.063(4)$\\ionosphere& $0.150(2)$& $0.163(6)$& $0.180(8)$& $0.149(1)$& $0.161(5)$& $0.156(4)$& $0.167(7)$& $0.155(3)$\\dermatology& $0.105(1)$& $0.178(5)$& $0.183(6)$& $0.188(7)$& $0.128(2)$& $0.143(3)$& $0.161(4)$& $0.214(8)$\\soybeansmallEW& $0.042(2)$& $0.064(6)$& $0.063(5)$& $0.035(1)$& $0.042(3)$& $0.049(4)$& $0.107(8)$& $0.065(7)$\\krvskpEW& $0.071(1)$& $0.086(3)$& $0.151(7)$& $0.094(6)$& $0.076(2)$& $0.088(4)$& $0.209(8)$& $0.092(5)$\\sonar& $0.221(1)$& $0.228(3)$& $0.223(2)$& $0.233(7)$& $0.231(5)$& $0.230(4)$& $0.312(8)$& $0.232(6)$\\libras& $0.402(2)$& $0.525(5)$& $0.499(4)$& $0.572(6)$& $0.400(1)$& $0.457(3)$& $0.572(7)$& $0.603(8)$\\semeionEW& $0.308(1)$& $0.531(6)$& $0.443(4)$& $0.466(5)$& $0.359(2)$& $0.425(3)$& $0.537(7)$& $0.641(8)$\\arrhythmia& $0.395(1)$& $0.428(5)$& $0.487(8)$& $0.423(3)$& $0.397(2)$& $0.424(4)$& $0.433(6)$& $0.486(7)$\\ujiIndoor& $0.052(2)$& $0.465(8)$& $0.381(5)$& $0.153(4)$& $0.124(3)$& $0.052(1)$& $0.458(6)$& $0.458(7)$\\pcmac& $0.333(1)$& $0.420(7)$& $0.400(6)$& $0.348(3)$& $0.370(5)$& $0.338(2)$& $0.494(8)$& $0.349(4)$\\basehock& $0.316(1)$& $0.423(6)$& $0.441(7)$& $0.343(2)$& $0.387(5)$& $0.343(3)$& $0.485(8)$& $0.343(4)$\\gisette& $0.068(2)$& $0.069(3)$& $0.252(7)$& $0.098(4)$& $0.160(6)$& $0.064(1)$& $0.311(8)$& $0.098(5)$\\pengleukEW& $0.056(3)$& $0.061(5)$& $0.319(8)$& $0.051(2)$& $0.060(4)$& $0.143(7)$& $0.076(6)$& $0.047(1)$\\gli-85& $0.175(3)$& $0.184(4)$& $0.195(5)$& $0.166(1)$& $0.231(7)$& $0.230(6)$& $0.308(8)$& $0.166(2)$\\
 \hline
Average rank & $1.6$& $5.0$& $5.8$& $4.0$& $4.0$& $3.4$& $7.2$& $5.2$\\
\bottomrule
\end{tabular}
\end{table*}
for the KNN classifier, and the Table \ref{tab.svmlin_results_allfilter}
\begin{table*}[!t]
\caption{Average misclassification error over $30$ repetitions and the rank of the FS methods for the SVMlin classifier.}
\centering
\label{tab.svmlin_results_allfilter}
\tiny
\setlength{\tabcolsep}{0.7 pt}
\begin{tabular}{|l|c|c|c|c|c|c|c|c|}
\toprule
& \rot{\textbf{HOCMIM}} & \rot{\textbf{RELIEFF}} & \rot{\textbf{ILFS}} & \rot{\textbf{Fischer}} & \rot{\textbf{NCA}} & \rot{\textbf{FS-OLS}}  & \rot{\textbf{Kruska}}  & \rot{\textbf{GINI}}   \\ 
\hline
parkinsonsEW& $0.337(7)$& $0.335(6)$& $0.326(3)$& $0.324(1)$& $0.339(8)$& $0.331(5)$& $0.330(4)$& $0.325(2)$\\spectEW& $0.377(1)$& $0.380(4)$& $0.380(5)$& $0.378(2)$& $0.392(8)$& $0.383(6)$& $0.387(7)$& $0.378(3)$\\german& $0.325(2)$& $0.329(5)$& $0.349(7)$& $0.326(4)$& $0.333(6)$& $0.324(1)$& $0.361(8)$& $0.325(3)$\\breastEW& $0.054(1)$& $0.063(6)$& $0.065(7)$& $0.061(3)$& $0.061(4)$& $0.056(2)$& $0.067(8)$& $0.061(5)$\\wdbc& $0.043(3)$& $0.045(4)$& $0.052(7)$& $0.048(5)$& $0.043(2)$& $0.043(1)$& $0.157(8)$& $0.048(6)$\\ionosphere& $0.180(5)$& $0.177(4)$& $0.218(8)$& $0.169(3)$& $0.215(7)$& $0.151(1)$& $0.187(6)$& $0.163(2)$\\dermatology& $0.095(1)$& $0.159(5)$& $0.164(6)$& $0.167(7)$& $0.119(2)$& $0.123(3)$& $0.137(4)$& $0.171(8)$\\soybeansmallEW& $0.038(4)$& $0.043(5)$& $0.044(7)$& $0.035(3)$& $0.034(2)$& $0.026(1)$& $0.071(8)$& $0.044(6)$\\krvskpEW& $0.066(2)$& $0.066(3)$& $0.116(7)$& $0.070(5)$& $0.076(6)$& $0.063(1)$& $0.188(8)$& $0.070(4)$\\sonar& $0.298(5)$& $0.285(1)$& $0.295(3)$& $0.299(6)$& $0.294(2)$& $0.303(7)$& $0.352(8)$& $0.296(4)$\\libras& $0.545(1)$& $0.653(5)$& $0.619(4)$& $0.671(7)$& $0.552(2)$& $0.579(3)$& $0.668(6)$& $0.699(8)$\\semeionEW& $0.313(1)$& $0.546(6)$& $0.448(4)$& $0.476(5)$& $0.373(2)$& $0.424(3)$& $0.569(7)$& $0.644(8)$\\arrhythmia& $0.409(1)$& $0.475(3)$& $0.494(5)$& $0.486(4)$& $0.438(2)$& $0.499(6)$& $0.515(8)$& $0.499(7)$\\penglungEW& $0.423(2)$& $0.512(7)$& $0.468(6)$& $0.463(4)$& $0.409(1)$& $0.464(5)$& $0.525(8)$& $0.449(3)$\\ujiIndoor& $0.048(1)$& $0.271(6)$& $0.188(5)$& $0.128(4)$& $0.105(3)$& $0.055(2)$& $0.296(7)$& $0.296(8)$\\pcmac& $0.337(2)$& $0.383(7)$& $0.362(6)$& $0.344(4)$& $0.352(5)$& $0.326(1)$& $0.473(8)$& $0.344(3)$\\basehock& $0.310(1)$& $0.383(7)$& $0.340(6)$& $0.333(3)$& $0.339(5)$& $0.311(2)$& $0.471(8)$& $0.333(4)$\\gisette& $0.110(2)$& $0.118(5)$& $0.251(7)$& $0.116(4)$& $0.162(6)$& $0.062(1)$& $0.294(8)$& $0.116(3)$\\pengleukEW& $0.065(1)$& $0.069(3)$& $0.377(8)$& $0.067(2)$& $0.093(5)$& $0.142(6)$& $0.181(7)$& $0.070(4)$\\gli-85& $0.237(1)$& $0.239(3)$& $0.296(7)$& $0.239(2)$& $0.274(5)$& $0.287(6)$& $0.398(8)$& $0.243(4)$\\
 \hline
Average rank & $2.2$& $4.8$& $5.9$& $3.9$& $4.2$& $3.1$& $7.2$& $4.8$\\
\bottomrule
\end{tabular}
\end{table*}
for the SVM-lin classifier. The results in the Tables \ref{tab.knn_results_allfilter} and \ref{tab.svmlin_results_allfilter} indicates that the HOCMIM outperforms in the majority of the filter FS algorithms in all datasets. The HOCMIM has an average score of 1.6 for the KNN, and 2.2 for the SVM-lin classifier. In complement to the Tables, the critical diagram Fig. \ref{fig.cd_allfilter},
\begin{figure*}[!t]
\centering
\subfigure[] 
{\label{fig.cd_knn_allfilter}\includegraphics[width=.49\columnwidth]{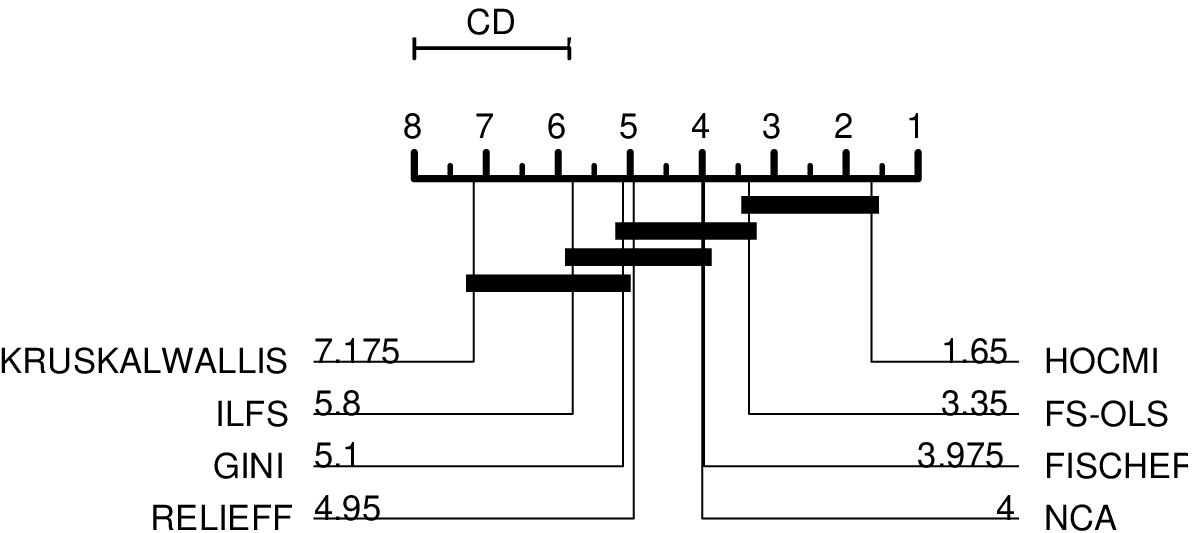}}
\subfigure[]
{\label{fig.cd_svmlin_allfilter}\includegraphics[width=.485\columnwidth]{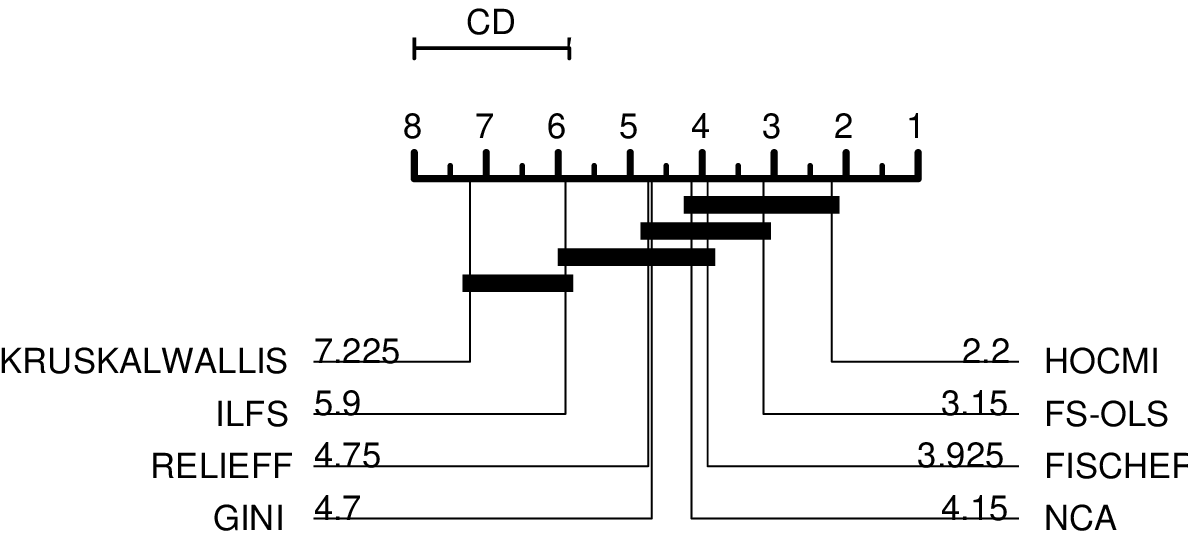}}
\caption{CD on the UCI benchmarking datasets for all filter methods, using the a) KNN classifier, b) SVM-lin classifier.}
\label{fig.cd_allfilter}
\end{figure*}\relax
shows that for the KNN classifier (Fig. \ref{fig.cd_knn_allfilter}), the HOCMIM outperforms all other methods with statistically significance, with except the FS-OLS algorithm. The Kruskal-Wallis test performs worse, followed by the ILFS. For the SVM-lin classifier (Fig. \ref{fig.cd_svmlin_allfilter}) the HOCMIM is statistically similar to the FS-OLS, Fischer, and NCA. Similarly, to the KNN, the Kruskal-Wallis test performs worse, followed by the ILFS. Also, from the results, it is noticeable the performance of the Fischer test, which stands in third for both classifiers, with an average rank of 4.0 for the KNN, and 3.9 for the SVM-lin classifier.

The line charts indicating the average rank over the all datasets datasets are indicated in Fig. \ref{fig.error_plot_average_all_filter}.
\begin{figure*}[!t]
\centering
\subfigure[]
{\label{fig.knn_error_plot_average_all_filter}\includegraphics[width=0.49\columnwidth]{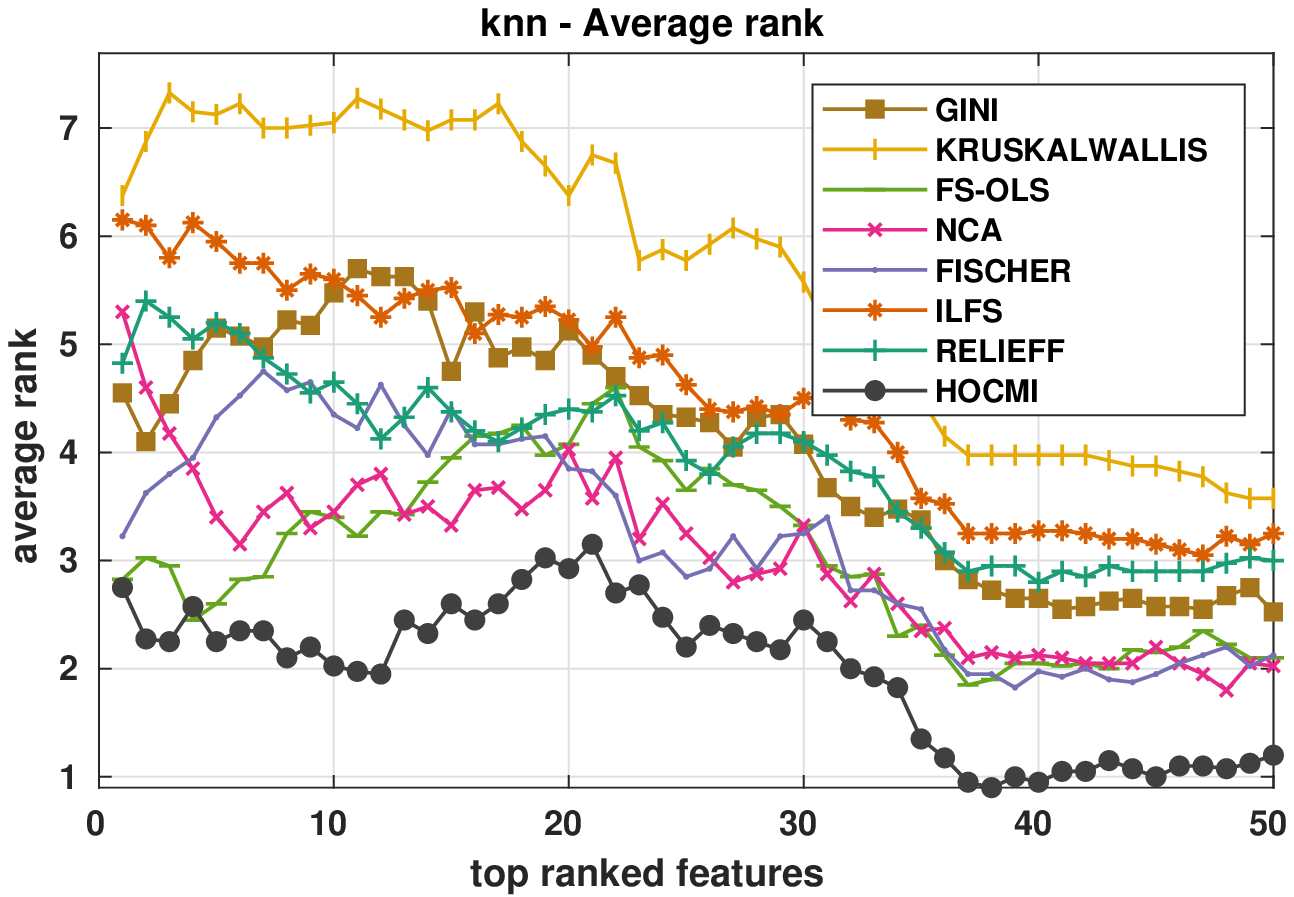}}
\subfigure[]
{\label{fig.svmlin_error_plot_average_all_filter}\includegraphics[width=0.49\columnwidth]{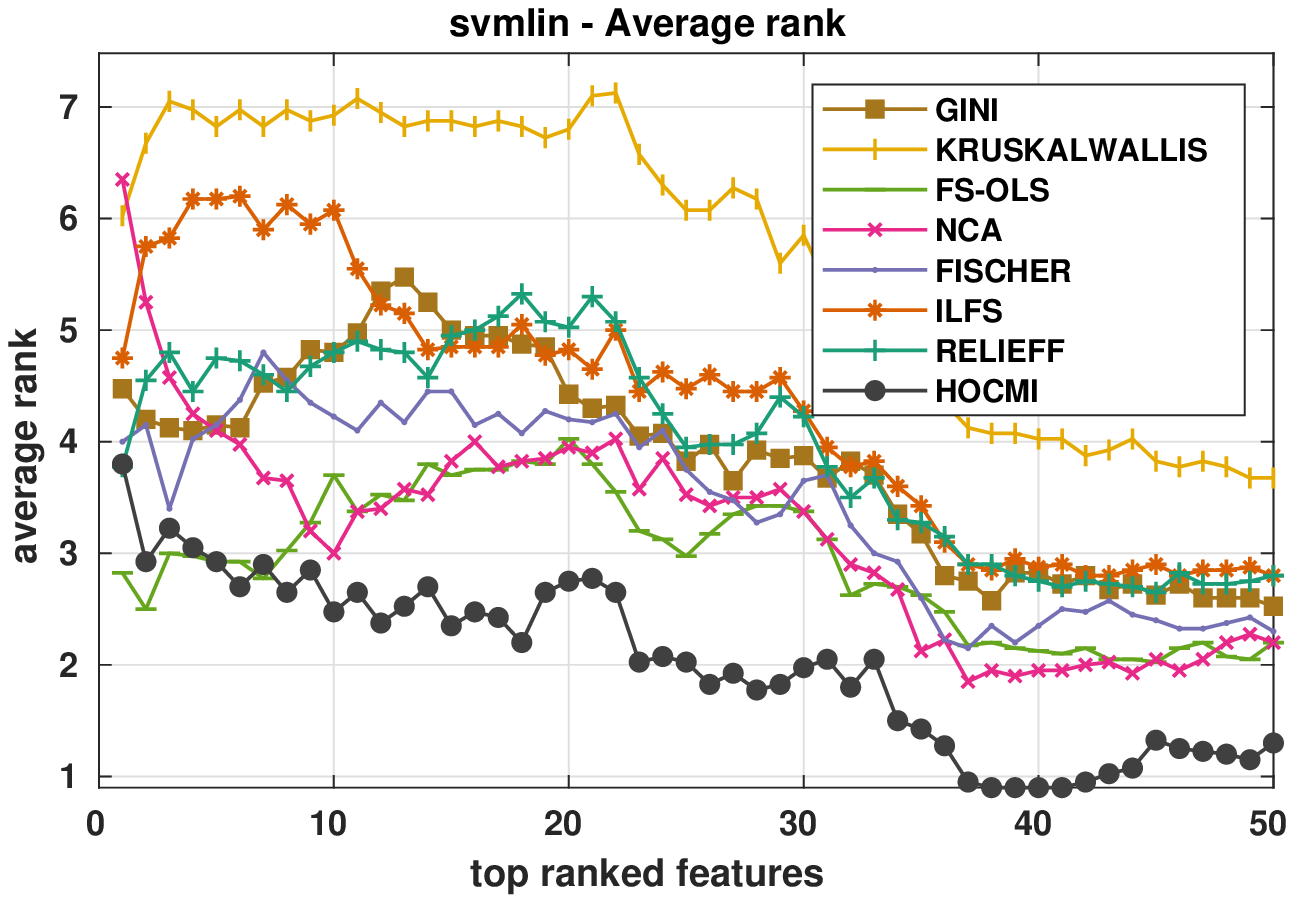}}
\caption{Top ranked features vs the rank of the HOCMIM and state-of-art filter FS algorithms (averaged over all datasets) for a) KNN and b) SVMlin classifiers.}
\label{fig.error_plot_average_all_filter}
\end{figure*}\relax
The results show that the HOCMIM stands as the lowest ran for both classifiers, indicating a clear better performance than the state-of-art filter FS algorithms.  

}


\subsection{Execution time}
{\color{black}The average and standard deviation of all method execution times for the datasets presented in Section \ref{subsec_bench_results}} are depicted in Fig. \ref{fig.exetime1}.
\begin{figure}[!t]
\centering
\includegraphics[width=.49\columnwidth]{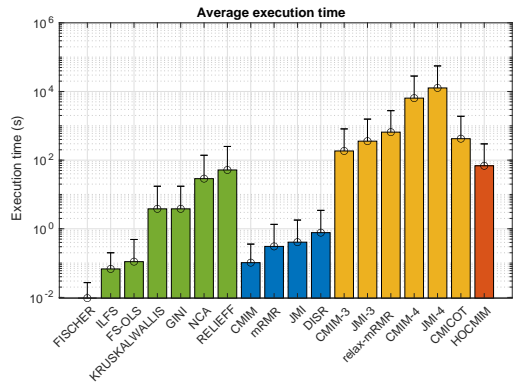}
\caption{Average execution time for all the FS algorithms.}
\label{fig.exetime1}
\end{figure}\relax
{\color{black} The lines above the bars indicates the standard deviation.}
To make the comparison fair, all algorithms which did not have a Matlab version were re-implemented for computing the execution time. 

{\color{black} The filter FS algorithms (green color) are ranked from fastest (left) to slowest (right).}
The {\color{black}MI FS} algorithms are ordered according to their order of dependency. {\color{black} From the execution time, the Fischer algorithm is the fastest  while the JMI-4 is the slowest. Apart from the Ficher, ILFS, and FS-OLS, all filter FS algorithms are slower in speed than the low order MI FS methods (CMIM, mRMR, JMI, and DISR)}. For the $3$rd order dependency, the relax-mRMR is the most computationally intensive, when compared to CMIM-3 and JMI-3. As expected, the CMIM-4, and JMI-4 are the most computationally `heavy'. {\color{black}The HOMCIM is slower than the filter FS algorithms and the low order MI FS methods}. On the other hand, the HOCMIM implementation demonstrated to be `light-weight' {\color{black} compared to high order MI FS algorithms}, being up to 6 times faster than CMICOT, 10 times faster than the relax-mRMR, and up to 100 faster than CMIM-4 and JMI-4 algorithms.



\section{Conclusion}
\label{sec.conclusion}


This paper proposed a new FS algorithm based on the mutual information (MI) theory, called high order conditional mutual information maximization (HOCMIM) and is based on the high order approximation of Conditional MI (CMI). The HOCMIM was derived from a `natural' expansion of CMI by formulating the method according to a maximization problem. A greedy search procedure was used to speed up the FS procedure by adaptively selecting the order $n$ and the best representative set.  {\color{black}The proposed HOCMIM framework opens up new perspectives on the exploration of high order dependencies in MI FS. For example, the HOCMIM framework can be used to understand the mechanisms underlying existing MI FS algorithms such as CMIM, CMIM-3, and CMIM-4, all of which have been explored for FS but are not fully understood.}

In the experimental part, HOCMIM was evaluated against {\color{black}eighteen state-of-the-art FS methods: Relieff, ILFS, Fischer, NCA, FS-OLS, Gini,} mRMR, DISR, CMIM, JMI, CMICOT, RelaxMRMR, JMI, DISR, CMIM-3, CMIM-4, JMI-3 and JMI-4, {\color{black} over twenty datasets from UCI repository}. The selected subsets were then evaluated using two different classifiers, the KNN and SVM-lin. {\color{black}The results indicates that the HOCMIM outperforms all the FS methods, mainly in the cases where where the ratio $N/D$ (number of examples per features) is higher in contrast to its counterparts. The HOCMIM still performs worse in small samples datasets. A possible cause is the accuracy of the MI estimator, which can fail to capture the high order dependencies with small number of samples.} Furthermore, the HOCMIM shows to be accurate and while allowing for high-order dependencies to be considered while maintaining a low computational cost when compared to higher-order counterparts.

{\color{black}Future research could look into how to improve the HOCMIM algorithm's execution time even more. Another factor that appears to be relevant is the use of HOCMIM with small sample datasets, which is still unresolved and must be addressed, either in terms of the MI estimator or by developing a new score function that takes small sampling into account. In addition, the derivation of of HOCMIM opens up new possibilities for high order dependencies MI FS algorithms.}

\bibliographystyle{IEEEtran}
\bibliography{pr2022_FranciscoSouza.bib}

%
%
%
%
%


\appendix
\section{HOCMIM toy example}
Consider the following example to better understand the concept underlying HOCMIM. The purpose in this example is to choose the first four most relevant features from the pool $\{X_1,X_2,X_3,X_4,X_5\}$ in relation to the target $Y$, where the following relation holds
\begin{equation}
    Y = (X_1\oplus X_2) \oplus  (X_3\oplus X_4) ,
\end{equation}
where, $\oplus$ is the xor operator, and feature $X_5$ is a irrelevant feature. The features and the output samples are listed in Table \ref{tab.artificial}.
\begin{table}[!t]
\caption{Toy example used to easily understand the HOCMIM.}
\label{tab.artificial}
{\scriptsize
\begin{center}
\setlength{\tabcolsep}{4 pt}
\begin{tabular}{c c c c c | c}
\toprule
$X_1$ & $X_2$  & $X_3$ & $X_4$ & $X_5$ &  $Y$ \\
\hline
     0   &  1 &    0    & 0&     1&     1\\
     1    & 1    & 1    & 1&     0     &0\\
     0    & 0   &  0    & 0&     0    & 0\\
     1    & 0  &   0&     0 &    0    & 1\\
     1    & 1 &    1 &    0  &   0    & 1\\
     0    & 0    & 0  &   1   &  0    & 1\\
     1    & 0   &  1   &  0  &   0   &  0\\
     1    & 0  &   1    & 0  &  0   &  0\\
     1    & 1 &    0&     1   &  0 &    1\\
     1    & 0&     0 &    0    & 1&     1\\
\bottomrule
\end{tabular}
\end{center} 
}
\end{table}\relax

The HOCMIM was run three times, by assuming a fixed order $n$, and setting $n=1,2,3$, and with features ranked accordingly. The ranking of the features for each run is shown Table \ref{tab.artrank}.
\begin{table}[!t]
\caption{Rank of HOCMIM selected variables}
\label{tab.artrank}
{\footnotesize
\begin{center}
\setlength{\tabcolsep}{4 pt}
\begin{tabular}{c c }
\toprule
$n$ & rank  \\
\hline
1 &  $\{X_3, X_2, X_4, X_5, X_1\}$ \\
2 &  $\{X_3, X_2, X_4, X_1, X_5\}$ \\
3 &  $\{X_3, X_2, X_4, X_1, X_5\}$ \\
\bottomrule
\end{tabular}
\end{center} 
}
\end{table}\relax
Because the irrelevant variable $X_5$ is ordered as more relevant than $X_
1$, the HOCMIM with order $n=1$ fails to rank the most relevant features correctly. As the order of HOCMIM increases, it tends to take into account the higher order interaction between the features and, as a result, correctly rank the features.

In the first iteration of HOCMIM algorithm, the MI of each pair $(X_i,Y)$ is computed. From the samples listed in Table \ref{tab.artificial}, the MI of each pair is computed as:
\begin{align}
    I(X_1,Y) &= 0.01\nonumber\\
    I(X_2,Y) &= 0.05\nonumber\\
    I(X_3,Y) &= 0.26\nonumber\\
    I(X_4,Y) &= 0.01\nonumber\\
    I(X_5,Y) &= 0.17\nonumber
\end{align}
Since, $I(X_3;Y)$ is the largest quantity, this is the first variable to be added to the pool of selected features $S=\{X_3\}$. To choose the next variable, the CMI between the remaining features and the output conditioned to $S$ is computed. The total redundancy of the unselected features is shown in Table \ref{tab.2totalR}. Because $S$ has only one feature, the total redundancy with order $n=1$ is the only one computed.
\begin{table}[!t]
\caption{$2$nd iteration of HOCMIM algorithm; $S=\{X_3\}$.}
\label{tab.2totalR}
{\footnotesize
\begin{center}
\setlength{\tabcolsep}{4 pt}
\begin{tabular}{c c c c c }
\toprule
$n$ &$R_n(X_1,S,Y), Z$ & $R_n(X_2,S,Y), Z$  & $R_n(X_4,S,Y), Z$ & $R_n(X_5,S,Y), Z$  \\
\hline
1 & -0.11, $\{X_3\}$ & -0.14, $\{X_3\}$ & -0.11, $\{X_3\}$ & 0.10, $\{X_3\}$\\
\bottomrule
\end{tabular}
\end{center} 
}
\end{table}\relax
The variable $X_2$ is selected as it scores the highest CMI:
\begin{align}
    I(X_2,Y|X_1) &= I(X_2;Y) - \max_{Z \cup S, |Z|=1} R(X_2,Z,Y) = 0.05 - (-0.14) = 0.19
\end{align}
The negative value in the total redundancy indicates that the conditional redundancy $I(X_2;X_3|Y)$ is greater than the redundancy $I(X_2;X_3)$, implying that $X_2$ contains more information when combined with $X_3$ than when used alone to predict $Y$. 

To select the third variable, the total redundancy is computed with order $n=1$, and $n=2$, with the set of selected features as $S=\{X_1,X_2\}$. The total redundancy and the respective representative set are listed in Table \ref{tab.3totalR}. 
\begin{table}[!t]
\caption{$3$rd iteration of HOCMIM algorithm; $S=\{X_2,X_3\}$.}
\label{tab.3totalR}
{\footnotesize
\begin{center}
\setlength{\tabcolsep}{4 pt}
\begin{tabular}{c c c c c }
\toprule
$n$ & $R_n(X_1,S,Y), Z$  & $R_n(X_4,S,Y), Z$ & $R_n(X_5,S,Y), Z$  \\
\hline
1 & -0.04, $\{X_2\}$ & -0.11, $\{X_3\}$ & 0.10, $\{X_3\}$ \\
2 & -0.11, $\{X_2,X_3\}$ & -0.24, $\{X_2,X_3\}$ & 0.12, $\{X_2,X_3\}$ \\
\bottomrule
\end{tabular}
\end{center} 
}
\end{table}\relax
In both, the total redundancy of the variables that scored highest according to     the HOCMIM criteria for $n=1$ and $n=2$ is variable $X_4$:
\begin{align}
    I_1(X_4,Y|S) &= I(X_4;Y) - \max_{Z \cup S, |Z|=1} R(X_4,Z,Y) = 0.01 - (-0.11) = 0.12\\
    I_2(X_4,Y|S) &= I(X_4;Y) - \max_{Z \cup S, |Z|=2} R(X_4,Z,Y) = 0.01 - (-0.24) = 0.25
\end{align}
Since $S=\{X_2,X_3\}$, the true total redundancy occurs when $n=2$. The total redundancy differs when $n=1$ and $n=2$, as shown in Table \ref{tab.3totalR}. When considering low order dependencies among features, $n=1$ can result in a significant error in determining true redundancy.

In the fourth stage, when approximating total redundancy with $n=1$, Table \ref{tab.4totalR}, 
\begin{table}[!t]
\caption{$4$rd iteration of HOCMIM algorithm; $S=\{X_1,X_2,X_4\}$.}
\label{tab.4totalR}
{\footnotesize
\begin{center}
\setlength{\tabcolsep}{4 pt}
\begin{tabular}{c c c c }
\toprule
$n$ & $R_n(X_1,S,Y), Z$ & $R_n(X_5,S,Y), Z$  \\
\hline
1 & -0.04, $\{X_2\}$ & 0.10, $\{X_3\}$  \\
2 & -0.08, $\{X_2,X_4\}$ & 0.12, $\{X_2,X_3\}$   \\
3 & -0.26, $\{X_1,X_2,X_3\}$ & 0.010, $\{X_2,X_3,X_4\}$  \\
\bottomrule
\end{tabular}
\end{center} 
}
\end{table}\relax
the variable to be chosen is $X_5$, with feature $Z=\{X_3\}$ picked as the representative set:
\begin{align}
    I_1(X_5,Y|S) &= I(X_5;Y) - \max_{Z \cup S, |Z|=1} R(X_5,Z,Y) = 0.17 - (0.10) = 0.07\\
    I_1(X_1,Y|S) &= I(X_1;Y) - \max_{Z \cup S, |Z|=1} R(X_1,Z,Y) = 0.01 - (0.04) = 0.05\\
    I_2(X_5,Y|S) &= I(X_5;Y) - \max_{Z \cup S, |Z|=1} R(X_5,Z,Y) = 0.17 - (0.12) = 0.05\\
    I_2(X_1,Y|S) &= I(X_1;Y) - \max_{Z \cup S, |Z|=1} R(X_1,Z,Y) = 0.01 - (0.08) = 0.09
\end{align}
It turns out that the approximation of the total redundancy of feature $X_1$, with $n=1$, does not approximate adequately the true total redundancy $R_3(X_1,S,Y)$. The variable $X_5$ is chosen when $n=2$, as the total redundancy approximation approaches the true value. From Table \ref{tab.4totalR}, the different of the true redundancy with a $n=1$, $n=2$, and $n=3$ are considerable. Using high order dependencies, which is the case of HOCMIM, benefits the FS and may reduce the error in approximating the true redundancy.

This example demonstrates how important it is to consider high order interactions between features. It also demonstrates that approximation by the HOCMIM criteria is a viable option, because it takes feature iteration into account in the selection procedure.

\end{document}